\documentclass[runningheads]{llncs}

\usepackage{eccv}

\usepackage{eccvabbrv}

\usepackage{graphicx}
\usepackage{booktabs}
\usepackage{multirow}
\usepackage{amssymb}
\usepackage{pifont}  
\usepackage{makecell}
\usepackage{tabularx}
\usepackage{enumitem}
\usepackage{wrapfig}
\usepackage{rotating}

\newcommand{\tit}[1]{\smallbreak\noindent\textbf{#1.}}

\def \ie {\emph{i.e.},}
\def \eg {\emph{e.g.},}

\def \wrt {\emph{w.r.t.}}

\usepackage[accsupp]{axessibility}

\usepackage{hyperref}

\usepackage{orcidlink}

\begin{document}

\title{Editing Everything Everywhere All at Once}

\titlerunning{Editing Everything Everywhere All at Once}

\author{Fabio Quattrini\thanks{Equal contribution, order determined by a coin flip.}\inst{1}\orcidlink{0009-0004-3244-6186} \and
Carmine Zaccagnino$^{\star}$\inst{1}\orcidlink{0009-0004-4953-6348} \and
Enis Simsar\inst{2}\orcidlink{0000-0002-6662-3249}  \and
Marta Tintoré Gazulla\inst{3} \and
Rita Cucchiara\inst{1}\orcidlink{0000-0002-2239-283X} \and
Alessio Tonioni\inst{3}\orcidlink{0000-0003-3358-9686} \and
Silvia Cascianelli\inst{1}\orcidlink{0000-0001-7885-6050}}

\authorrunning{F.~Quattrini, C. Zaccagnino et al.}

\institute{University of Modena and Reggio Emilia \and ETH Zurich \and Google}

\maketitle
\vspace{-2.5pt}
{\centering
\url{https://mice.silviacascianelli.com}\par
}
\vspace{-2.5pt}

\begin{abstract}
  Editing multiple elements of an image in a single forward pass is a practical alternative to multi-turn image manipulation, offering improved efficiency and potentially better harmonization. However, when several instructions target different regions, semantic interference often leads to attribute leakage and poor edit disentanglement, especially as the number of edits increases. 
  In this work, we propose \textbf{MICE} (Multi-Instance Concurrent Editing), a training-free strategy for scalable multi-instance image editing with Multimodal Diffusion Transformers. MICE modifies the additive bias of joint attention to regulate interactions between instance-specific edit instructions, latent, and context tokens identified via user-provided segmentation masks. Specifically, MICE allows intra-instance attention, penalizes interactions between neighboring region tokens, and suppresses unrelated cross-instance attention. As a result, our method enforces attribute binding while preserving global visual consistency. 
  We evaluate MICE on LoMOE-Bench and introduce \textbf{MICE-Bench}, a more challenging benchmark with an average of 8.5 concurrent edits per image. The experiments demonstrate that our approach outperforms strong baselines and recent competitors in terms of visual quality preservation and faithfulness to the editing instructions. 
  \keywords{Multi-Instance Editing \and Image Editing \and Flow Matching}
  \begin{figure}
    \centering
    \vspace{-0.2cm}
    \includegraphics[width=.99\linewidth]{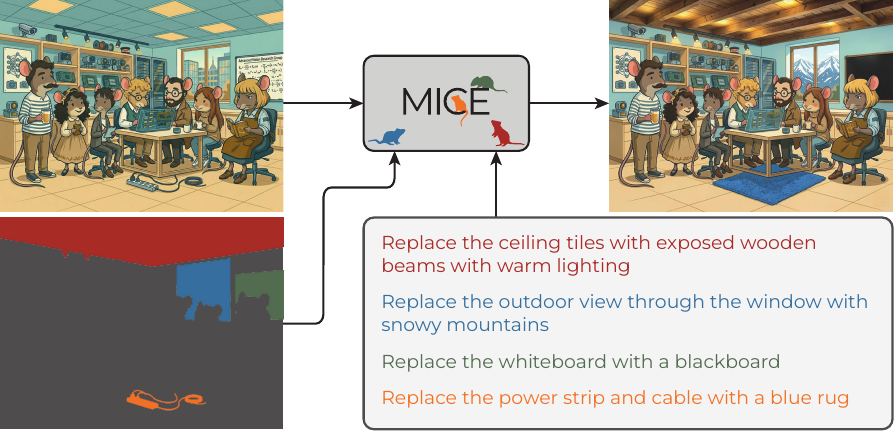}
    \vspace{-0.2cm}
    \caption{Our MICE approach modifies the joint attention maps of MMDiT-based flow matching generative models to concurrently edit multiple elements at inference time. }\vspace{-0.5cm}
    \label{fig:overview}
  \end{figure}
\end{abstract}

\section{Introduction}

Concurrent modification of multiple elements within a single image has recently emerged as a compelling extension of text-guided image editing~\cite{goirik_lomoe, fu2025layeredit, zaccagnino2026shifting, bar2023multidiffusion}. A possible solution would entail multi-turn pipelines~\cite{zhou2025multi, ye2025imgedit}, where edits are applied sequentially. However, editing multiple instances one at a time is time-consuming, and the final output can depend on the order in which the edits are performed in case of overlapping instances.
Concurrent editing enables all transformations to be performed in one pass, offering clear advantages in efficiency and usability. Nonetheless, existing approaches implementing this setting~\cite{labs2025flux_kontext,wu2025qwen_image} face a fundamental challenge: when several instructions target different regions or objects, their semantic signals may interfere, leading to \emph{attribute leakage} across instances. As the number of edits increases, ensuring edit disentanglement, spatial locality, and global visual coherence becomes progressively more difficult (a quantitative analysis of this phenomenon is reported in the supplementary).

To address this problem, existing approaches primarily entail inference-time adaptations of diffusion and flow-based generative models. A first family of methods performs gradient descent on the image latents to progressively correct undesired interactions between edits~\cite{goirik_lomoe, fu2025layeredit}. A second line of work intervenes directly in the attention mechanisms of Multimodal Diffusion Transformers~\cite{esser2024scaling}(MMDiTs), introducing biasing strategies at selected layers of the velocity field estimator to regulate token interactions~\cite{zaccagnino2026shifting, zhou2025dreamrenderer, qu2025replan}. Compared to gradient descent, attention-based solutions are generally more efficient and more readily applicable to transformer-based backbones. However, prior biasing strategies~\cite{zaccagnino2026shifting,zhou2025dreamrenderer} typically require defining multiple bias matrices to be applied to architecture-dependent layers determined by hyperparameters (e.g., a strict bias to isolate edits in certain layers and a softer bias to promote harmonization in other layers). This makes performance sensitive to model-specific design choices and reduces portability across architectures.

In this work, we propose a training-free and architecture-agnostic strategy for multi-instance concurrent editing, dubbed \textbf{MICE}. Our approach operates directly on the joint attention mechanism of MMDiTs by modifying the additive bias applied to the attention logits (which is commonly set to 0). Rather than enforcing hard structural constraints, we reshape token interactions in a principled manner to balance instance isolation with global coherence. Specifically, we modify the additive bias in the joint attention operation so that the interactions among prompt, latent, and context tokens relative to each edit instance are isolated from those of other edit instances, ensuring attribute binding, while still maintaining visibility to relevant out-of-instance tokens to enhance harmonization. 
We identify the instances via segmentation masks that can be provided by the user or the output of a promptable segmentation model~\cite{kirillov2023segment,wu2024general}, and use such masks to identify instance-specific tokens among the text prompt, latent, and context source image tokens given as input to the MMDiT. Then we compute the additive bias of the joint attention map using the masks, smoothed with a parametric Gaussian kernel. The effect of our strategy is that the tokens relative to a specific instance attend to each other with no bias, can attend to neighboring tokens with a decreasing negative bias (accommodating for small shape variations and helping harmonize the overall appearance of the image), and are prevented from attending faraway, non-relevant tokens (possibly relative to different editing instances).%
As a result, our approach prevents attribute leakage between editing instances while obtaining a harmonized, visually appealing overall image.

We instantiate our method on Flow Matching-based backbones and evaluate it on LoMOE-Bench~\cite{goirik_lomoe}, a recent benchmark designed for parallel multi-instance editing. Note that the samples in LoMOE-Bench typically contain 3 concurrent edits, rendering it insufficient for evaluating performance under challenging conditions. Therefore, to evaluate performance under more complex conditions, we introduce \textbf{MICE-Bench}, a new benchmark featuring samples with approximately 8.5 concurrent edits on average.\footnote{\url{https://hf.co/datasets/blowing-up-groundhogs/mice_bench}}
Across both benchmarks, MICE demonstrates superior instruction-following capability compared to strong baselines and recent competitors, both in terms of the number of edits correctly executed, background preservation, and the faithfulness of each modification to its corresponding textual instruction. For reproducibility, we release the code.\footnote{\url{https://github.com/Blowing-Up-Groundhogs/mice}}
\section{Related Work}
\label{sec:related}
Conditional generative image editing methods have seen great progress and consolidated around Autoregression~\cite{var2024,controlvar2024, infinity2024,team2025nextstep} or Diffusion~\cite{liu2025step1x,wu2025qwen_image,tan2025ominicontrol,wang2025seededit, labs2025flux_kontext} approaches, with Flow Matching (FM)~\cite{lipman2023flow} emerging as a dominant framework for efficient, straight-path sampling~\cite{tong2024improving}. The conditioning signal is generally multimodal, comprising guidance~\cite{zheng2023guided}, text instructions~\cite{esser2024scaling}, and context images~\cite{contextar2025,labs2025flux_kontext, wu2025qwen_image}. However, these techniques rely on globally conditioned velocity fields and struggle with local, fine-grained localization. Recent works address this by using attention regularization~\cite{simsar2025lime,zhang2025eligen} or constrained cross-token interactions~\cite{dahary2024beyourself, paralleledits2024} but primarily target single-region or global edits and often fail to maintain coherence when applied to multiple disjoint regions simultaneously. While some methods investigate region-aware generation~\cite{chen2024region, eijkelboomcontrolled} and causal masking~\cite{he2024calmflow} to inject spatial control, most FM editors~\cite{labs2025flux_kontext, wu2025qwen_image} still rely on globally conditioned velocity fields and suffer attribute leakage between regions during multi-instance editing~\cite{mun2025ale, zhou2025dreamrenderer}, a limitation we address via our instance-disentangled attention mechanism. 
To address the aforementioned challenges of multi-instance editing, current approaches primarily cluster into three categories: scaling inference time (\textit{multi-turn}), scaling the architecture (\textit{multi-branch}), or regularizing the internal generation process.

\tit{Multi-Turn} 
State-of-the-art editing models are trained to reduce visual drift and show promise in iterative editing workflows~\cite{labs2025flux_kontext, wu2025qwen_image}, where edits are performed sequentially. While serving as a functional baseline, multi-turn editing differs fundamentally from concurrent editing. Technically, visual degradation accumulates as the number of iterations exceeds a few cycles (around six~\cite{labs2025flux_kontext}), and inference time scales linearly with the number of instructions, making this approach infeasible as instance edit counts increase. Methodologically, the final result is path-dependent, especially with overlapping regions. We therefore focus on \textit{concurrent} editing, where all changes occur in a single pass.

\tit{Multi-Branch}
To enforce disentanglement, some approaches split the generation process into multiple parallel branches, typically one per edit or region. MultiDiffusion~\cite{bar2023multidiffusion}, the seminal method in this paradigm, binds multiple generation paths by merging independent denoising steps into a least-squares optimal solution. LoMOE~\cite{goirik_lomoe} adopts this framework to apply localized influence on the N target regions using foreground masks, while other approaches further refine this by using regional branches~\cite{chen2024region}, performing independent inversion searches~\cite{yang2024object} or grouping edit "aspects" into specialized branches~\cite{paralleledits2024}. Finally, LayerEdit~\cite{fu2025layeredit} takes this paradigm to its extreme by decomposing the scene into $N+2$ independent layers.
While multiple generation paths allow instance isolation, these methods suffer from severe scaling bottlenecks in time and memory usage. Moreover, computational overhead is worsened by the latent optimization often required. For instance, LayerEdit and LoMOE rely on iterative gradient descent and perform a backward pass over N parallel branches. Furthermore, these methods rely on heuristic reassembly to smooth visual boundaries, which often fails to maintain global coherence in dense, overlapping scenes.

\tit{Regularization}
Another class of methods focuses on end-to-end, single-pass editing by regularizing the internal generation process. One prominent direction involves inference-time optimization. MDE-Edit~\cite{zhu2025mde} performs gradient descent on the latents at each denoising step to minimize specialized losses. While these methods maintain a single-pass memory footprint, they rely on aggregating $O(N)$ loss terms, introducing significant computational overhead and numerical instability during backpropagation, particularly when handling multiple instances of varying spatial scales, where large-scale objects can dominate the gradient signal and suppress smaller targets. Other approaches perform fine-tuning or train additional modules to target quantity perception~\cite{li2025moedit} or personalization~\cite{matsuda2024multi}. Closer to our work, some approaches tweak the attention operator. IDAttn~\cite{zaccagnino2026shifting} introduces binary disentanglement masks ($0,-\infty$) to enforce instance isolation. However, this approach relies on heuristics to balance different masking regimes in different model layers (early, middle, and late). This rigidity makes performance sensitive to model architecture and hinders porting to other baselines. In contrast, MICE offers a principled, architecture-agnostic solution by replacing hard binary masks with a parametric additive bias, enabling a soft, continuous transition between instance isolation and global harmonization without per-layer tuning or optimization overhead.

\section{MICE}
\label{sec:method}
\setlength{\belowdisplayskip}{3pt} \setlength{\belowdisplayshortskip}{3pt}
\setlength{\abovedisplayskip}{3pt} \setlength{\abovedisplayshortskip}{3pt}

In this section, we present our approach for multi-instance image editing. 
The task consists in modifying the content of multiple areas of an image (instances) according to user-provided editing instructions, while preserving its original visual style and layout consistency. 
Our approach builds upon a recent MMDiT-based flow matching model capable of text-conditioned image editing, which we extend with a novel biased joint attention mechanism. Unlike prior work that relies on strict hard masking strategies, which often result in harsh boundaries and poor coherence, our method introduces a continuous, spatially-smoothed attention bias formulation, as detailed below.

\subsection{Preliminaries}
Formally, let $I^{\text{ref}} \in \mathbb{R}^{H \times W \times 3}$ denote the input image and consider a set of $N$ segmentation masks $M = \{m_n\}_{n=1}^N$, each enclosing an object, and a set of corresponding editing instructions $T = \{t_n\}_{n=1}^N$, \eg~\textit{“Change the mouse to a black chick”}.
The masks can either be provided by a user or by auxiliary modules for Instance Segmentation.
The task entails generating an edited image $I^{\text{edit}}$ such that each instance within $m_n$ is replaced according to $t_n$, while maintaining the global style and visual coherence of $I^{\text{ref}}$. 

Our framework operates within a MMDiT~\cite{esser2024scaling, labs2025flux_kontext} architecture trained with conditional rectified flow matching~\cite{lipman2023flow, tong2024improving}. Specifically, given a target data sample $I^{\text{edit}}_1 \sim p_{\text{data}}$, a source sample $I^{\text{edit}}_0 \sim p_0$, where $p_0 \sim \mathcal{N}(0, \mathbf{I})$, and conditioning signals $\mathcal{C}=\{I^{\text{ref}}, T, M\}$, the model learns a time-dependent velocity field $v_\theta(I^{\text{edit}}_t, t, \mathcal{C})$ to denoise $I^{\text{edit}}_0$ into $I^{\text{edit}}_1$. 
The input image $I^\text{ref}$ is encoded by a Variational Autoencoder (VAE), which compresses it to $\mathbb{R}^{\frac{H}{d}\times \frac{W}{d}\times c}$ where $d$ is the compression ratio and $c$ is the number of channels of the VAE. The editing instructions $t_i$ are tokenized, padded to a fixed length $l$, and then passed through a text encoder, usually a language model like T5\cite{raffel2020exploring} or Qwen\cite{yang2025qwen3}, which returns a sequence of $l$ tokens encoding the information contained in the editing instruction. Then, these are fed to a velocity field estimator, alongside the noisy latents for $I^{\text{edit}}_t$. After a predefined number of estimation steps, the resulting latents are passed to a VAE decoder to obtain the final $I^{\text{edit}}_0$.

In state-of-the-art FM models, the velocity field estimator is an MMDiT, where tokens from various modalities (textual prompt, noisy visual latents, and conditioning context) are concatenated and processed via joint attention.
Let $\mathbf{Z} = \mathbf{Z}^\text{prompt} \parallel \mathbf{Z}^\text{latent} \parallel \mathbf{Z}^\text{context}$ denote the joint token sequence. From $\mathbf{Z}$ we compute the queries $\mathbf{Q}$, keys $\mathbf{K}$, and values $\mathbf{V}$ for the joint attention operation defined as:
\begin{equation*}
    \text{Attn}(\mathbf{Q}, \mathbf{K}, \mathbf{V}) = \text{Softmax}\left(\frac{\mathbf{Q}\mathbf{K}^T}{\sqrt{d}} + \mathbf{B}\right)\mathbf{V},
\end{equation*}
where $\mathbf{B} \in \mathbb{R}_{\leq0}^{|\mathbf{Z}| \times |\mathbf{Z}|}$ is an attention bias matrix, commonly set to 0.
Our approach defines the values of the bias matrix based on the types of tokens that interact to achieve editing isolation and preserve background and overall visual quality.

\subsection{Smoothly-Disentangled Attention}
\begin{figure}[t]
    \centering
    \begin{tabular}{ccccc}
         \includegraphics[width=0.1925\linewidth]{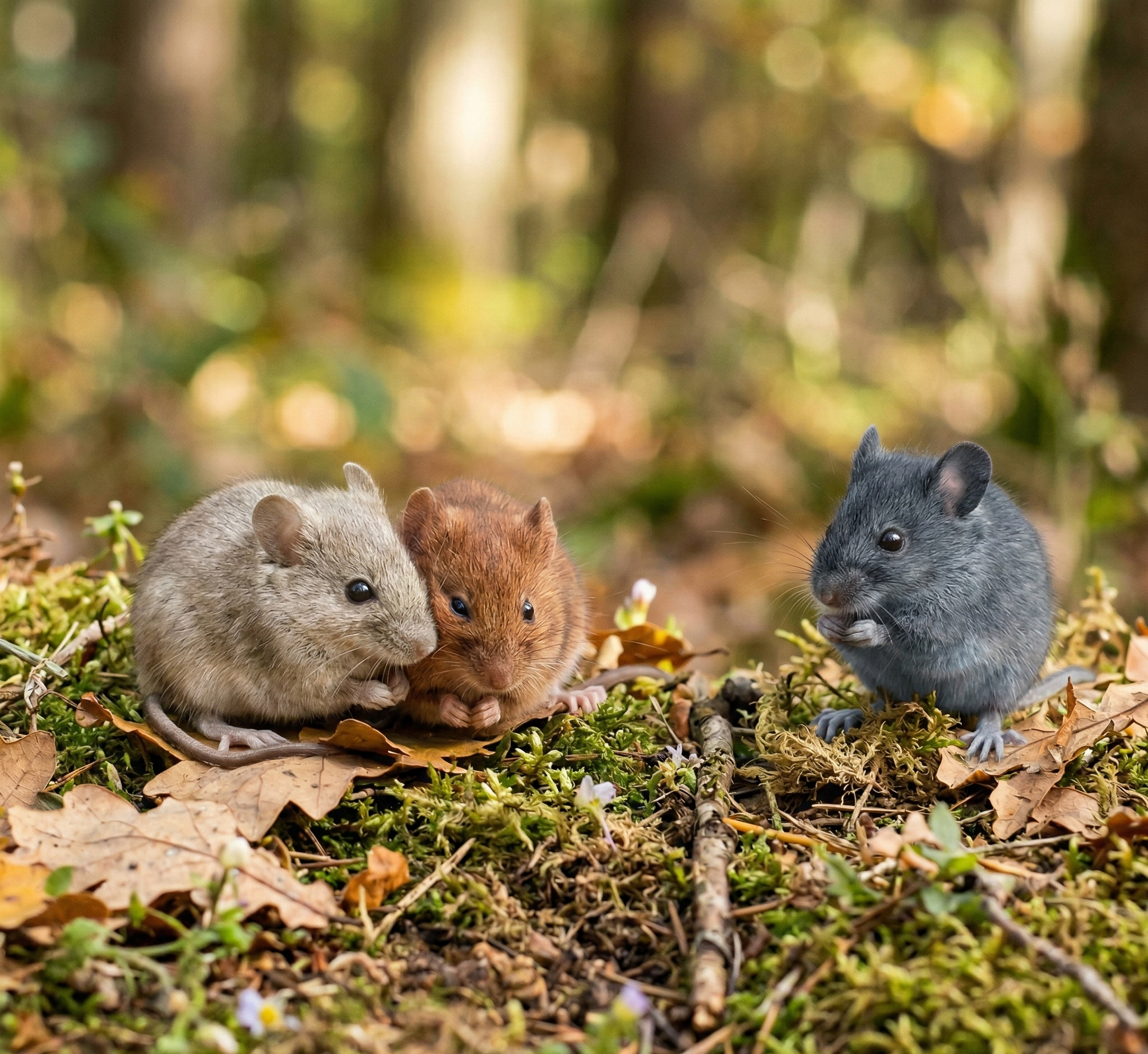} &
         \includegraphics[width=0.1925\linewidth]{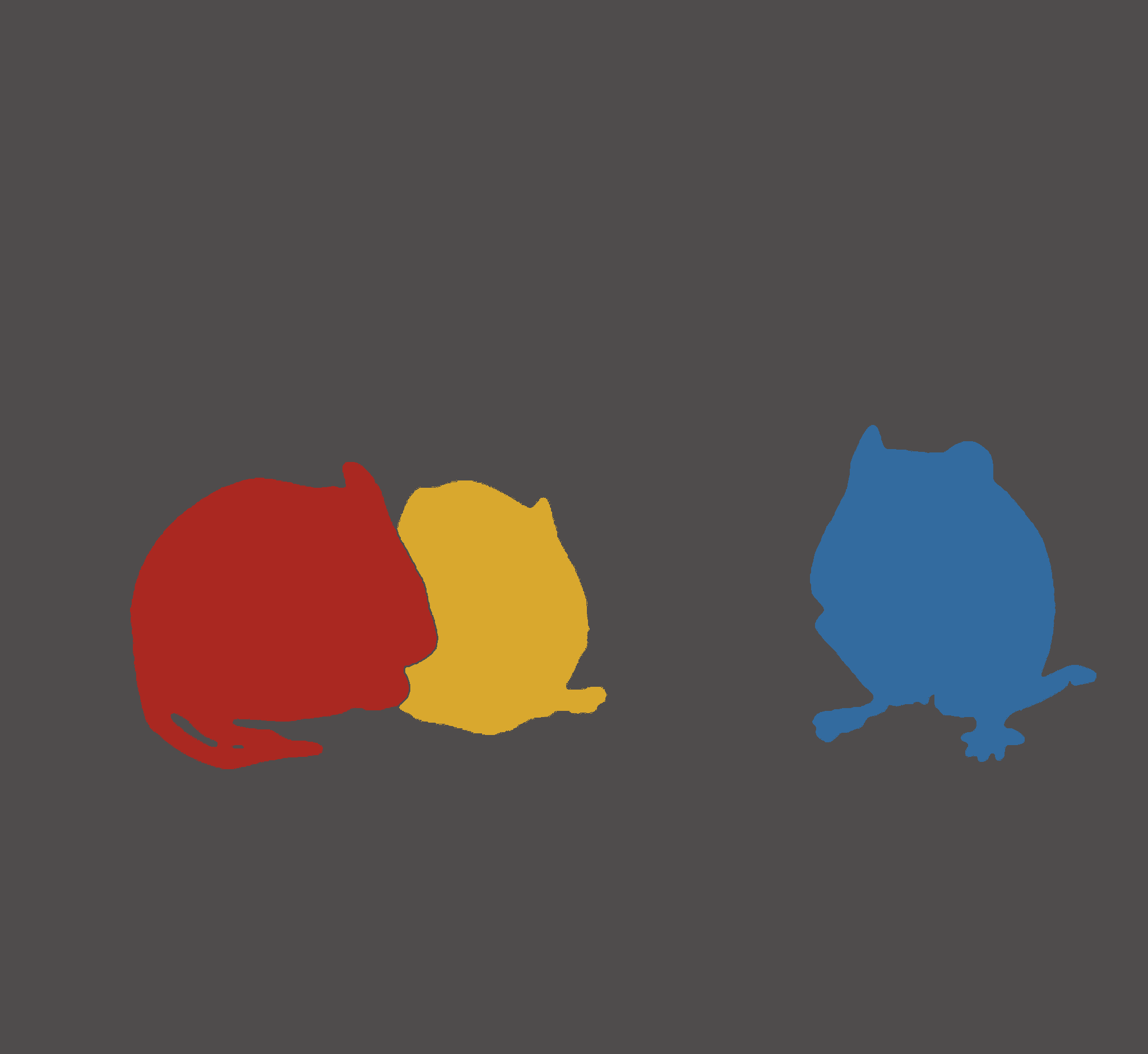} &
         \includegraphics[width=0.1925\linewidth]{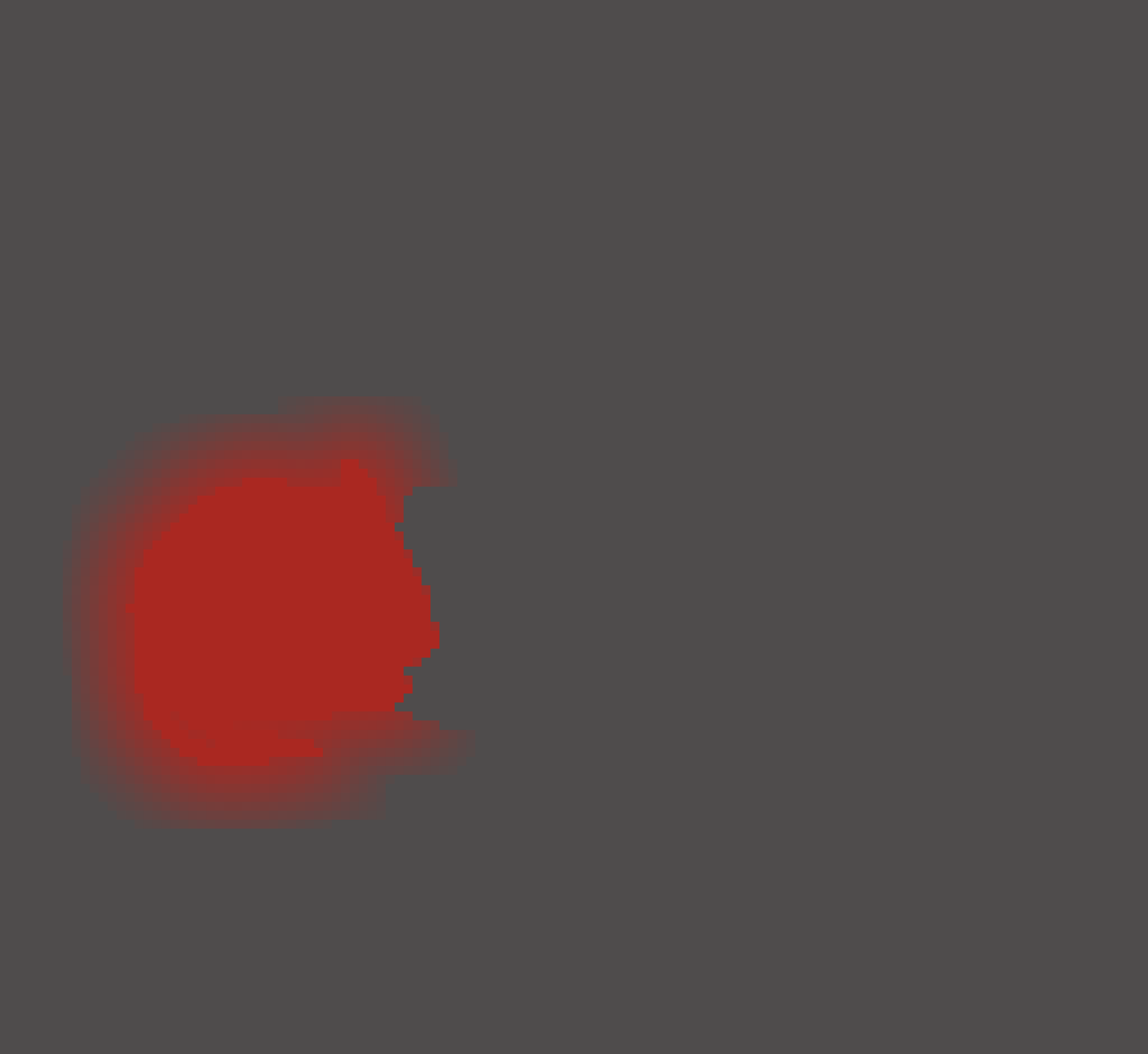} &
         \includegraphics[width=0.1925\linewidth]{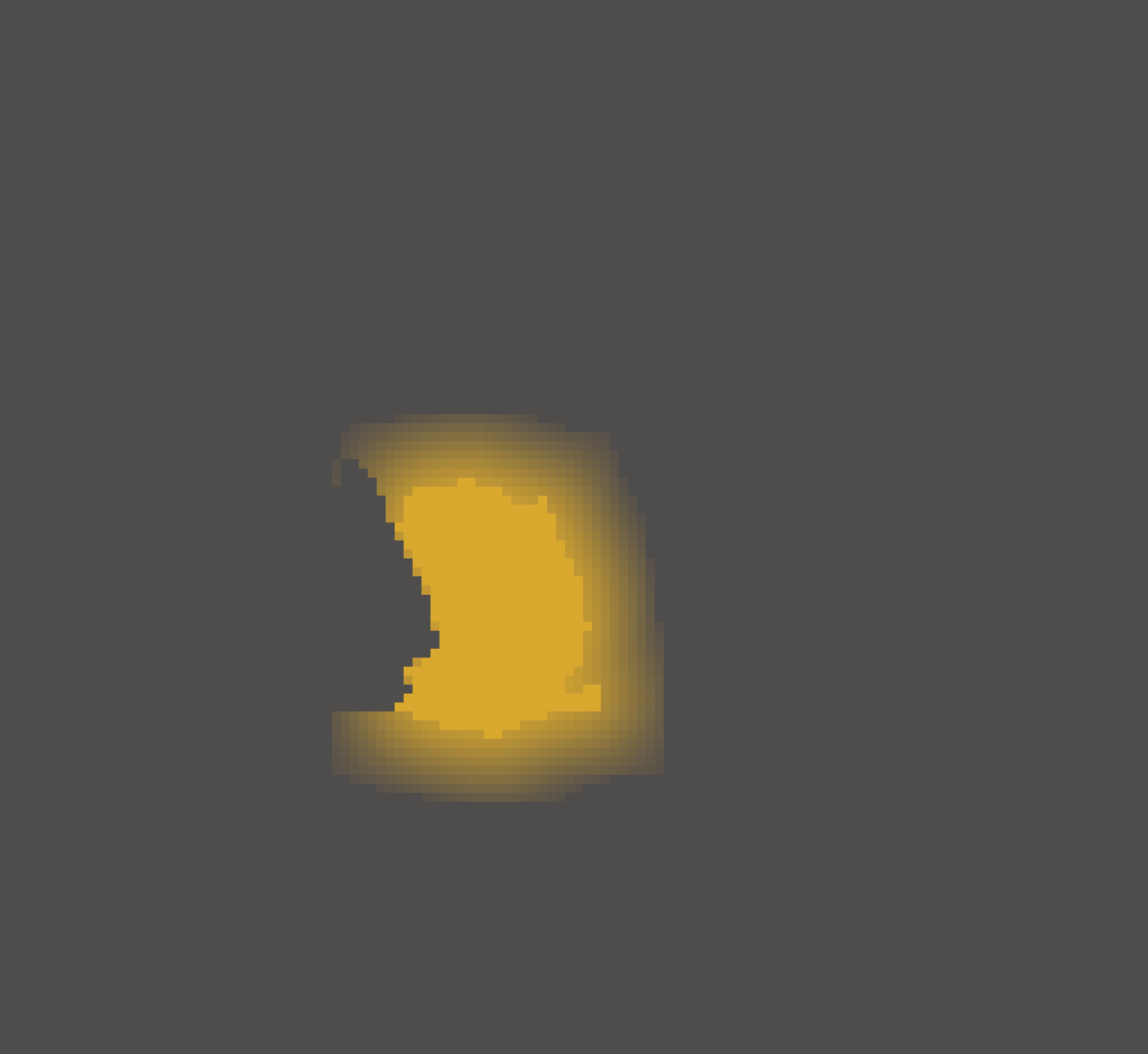} &
         \includegraphics[width=0.1925\linewidth]{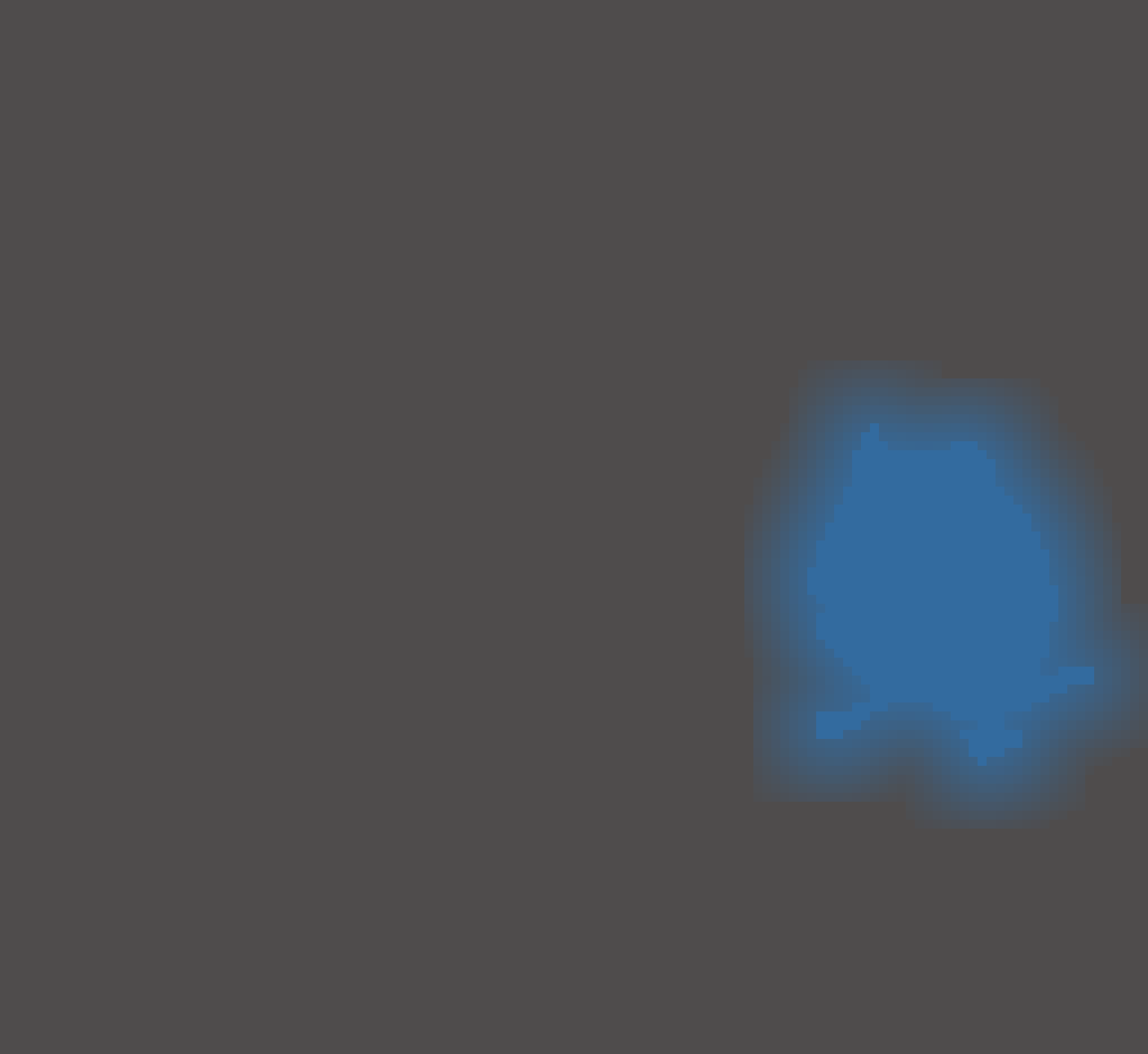}
    \end{tabular}
    \vspace{-0.35cm}
    \caption{Instance-aware smoothing strategy. Each segmentation mask is smoothed independently, and the smoothed values are zeroed when touching the edge of a neighboring, non-smoothed mask (\eg~the red and yellow masks). If an instance is sufficiently distant from the others, its mask is smoothed with no constraints (\eg~the blue mask).    \vspace{-.65cm}}
    \label{fig:smoothing}
\end{figure}

Prior instance-disentangled attention methods~\cite{zaccagnino2026shifting,zhou2025dreamrenderer,qu2025replan} enforce isolation by resorting to binary attention bias matrices $\mathbf{B}$, with $0$ allowing interactions and $-\infty$ blocking them. This approach is effective for semantic isolation but can lead to noticeable seams, incorrect local lighting, and a lack of smooth blending with the nearby areas of the image. To contrast this, some approaches~\cite{zhou2025dreamrenderer,zaccagnino2026shifting} relax disentanglement by allowing full latent and context self- and cross-interactions in some layers of the MMDiT. However, this strategy entails identifying the optimal layers where to apply the disentangled and relaxed regimes for each architecture, limiting the method portability.  
To tackle this, we propose to apply the same bias matrix in all layers by designing it to ensure a trade-off between disentanglement and harmonization. 
Specifically, we not only set the values of the matrix to $0$ to allow interactions and $-\infty$ to prevent them, but also use intermediate negative values to dampen the attention without fully blocking it. 

\tit{Editing Areas Definition}
To determine the values of the bias, we start from the segmentation masks of each instance, which can be either provided directly by the user or obtained from an instance segmentation model. We smooth such masks via a Gaussian kernel so that the attention will be increasingly penalized as the spatial distance from the instance mask increases.
Specifically, given a binary 2D segmentation mask $m_n \in \{0, 1\}^{H \times W}$ for instance $n$, we downsample it to account for VAE compression, then we apply a 2D Gaussian convolution with kernel size $k$ and obtain a smoothed mask $s_n$. 
To prevent interference between instances due to the smoothing, we devise a specific instance-aware smoothing strategy. We consider each instance separately and apply the Gaussian smoothing to its segmentation mask. Given instance $n$, if an element in the latent-space feature map is part of the mask of any other instance, its value is explicitly overridden to $0$ (see~\cref{fig:smoothing}).

The resulting continuous values lie in $[0,1]$, therefore, we map them into the log-probability space $(-\infty,0]$ to serve as an attention bias. 
Once the smoothed masks are defined and corrected to ensure instance-awareness, we unroll them in raster order. Each of them becomes a row in the bias matrix $\mathbf{B}$, whose values are defined as follows. 
Given a visual token $Z_j$ (either in $\mathbf{Z}^{\text{latent}}$ or $\mathbf{Z}^{\text{context}}$) outside the strict bounds of instance $n$, the bias in the attention with a latent or context token $Z_i$ belonging to instance $n$ is defined as:
\begin{equation*}
    \mathcal{B}(j) =  \frac{1}{\tau}\log(\mathbf{s}_{n}[j] + \epsilon)
\end{equation*}
where $\tau$ is a temperature hyperparameter controlling the falloff intensity, and $\epsilon=10^{-30}$ prevents numerical instability. 
Intuitively, $k$ and $\tau$ control the spatial extent and the strength of the blending between the edited instance and the surrounding regions. Large values for $k$ allow tokens to interact with a wider neighborhood, which is beneficial when the edit modifies the object shape or when the mask is irregular, and large values for $\tau$ typically yield smoother blending of the edited instance with the surrounding background. On the other hand, smaller values for these hyperparameters keep the editing more localized. A deeper analysis of the effect of $k$ and $\tau$ is reported in the supplementary material.

\begin{figure}[t]
    \centering
    \includegraphics[width=\linewidth]{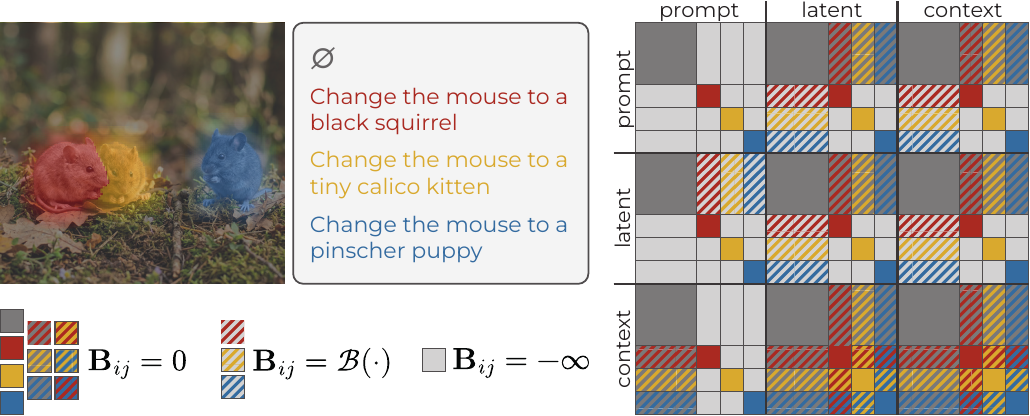}
    \vspace{-0.5cm}
    \caption{Our MICE approach entails defining a bias map (whose logic representation is on the right) that regulates the interaction between background and instance-specific prompt, latent, and context tokens, based on instances' localization information in the form of smoothed segmentation masks.}\vspace{-0.6cm}
    \label{fig:bias}
\end{figure}

The definition of the overall matrix is detailed below and schematized in~\cref{fig:bias}. 

\tit{Bias Matrix}
To define the bias matrix, we logically partition the sequence of tokens $\mathbf{Z}$ according to the $N$ instances and their corresponding target regions in the source context image and the generated image. Specifically:
\begin{itemize}[noitemsep, topsep=0pt,leftmargin=*]
    \item The \textbf{prompt tokens} $\mathbf{Z}^\text{prompt}$ are grouped in sequences corresponding to each editing prompt for the instances $\mathbf{Z}^\text{prompt}_n$ and an auxiliary sequence that represents the global prompt $\mathbf{Z}^\text{prompt}_g$. This can either be empty to keep the original overall appearance of the image, or instantiated to change its global attributes (\eg~style, lighting, color palette). 
    \item The \textbf{latent tokens} $\mathbf{Z}^\text{latent}$ are organized into background tokens $\mathbf{Z}^\text{latent}_u$ that should stay unedited, and local instance tokens $\mathbf{Z}^\text{latent}_n$ corresponding to the area that will be occupied by each edited instance in the final image.
    \item The \textbf{context tokens} $\mathbf{Z}^\text{context}$ are distinguished between tokens belonging to background $\mathbf{Z}^\text{context}_u$ and instances $\mathbf{Z}^\text{context}_n$ similarly to the latent tokens.
\end{itemize}

To formally align the global token sequence with the spatial dimensions of the masks, we define a mapping function that yields the spatial raster index of any visual token $Z_j \in \mathbf{Z}^{\text{latent}} \cup \mathbf{Z}^{\text{context}}$, \ie~$\pi(\cdot): \{0, ..., |\mathbf{Z}|\} \rightarrow \{ 0, ..., \frac{H}{d}\times\frac{W}{d}\}$. The attention bias matrix is thus defined as:
\begin{equation*}
\small
    \label{eq:mask_dis}
    \mathbf{B}_{ij} =  
    \left\{ \begin{array}{l l r}
    0 & (Z_i, Z_j) \in \mathbf{Z}^{\text{prompt}}_g \lor (Z_i, Z_j) \in \mathbf{Z}^{\text{prompt}}_n~\forall n \in [1, N] & \text{(1)} \\[6pt]
    0 & Z_i \in \mathbf{Z}^{\text{prompt}}_g \cup \mathbf{Z}^{\text{latent}}_u \land Z_j \notin \mathbf{Z}^{\text{prompt}} & \text{(2)} \\[6pt]
    0 & Z_i\in \mathbf{Z}^{\text{latent}}_g \cup \mathbf{Z}^{\text{context}} \land Z_j\in \mathbf{Z}^{\text{prompt}}_g & \text{(3)} \\[6pt]
    0 & Z_i \in \mathbf{Z}^{\text{context}}_n \land Z_j \in \mathbf{Z}^{\text{prompt}}_n~\forall n \in [1, N] & \text{(4)} \\[6pt]
    0 & Z_i \in \mathbf{Z}^{\text{context}} \land Z_j \notin \mathbf{Z}^{\text{prompt}} & \text{(5)} \\[6pt]
    \mathcal{B}(\pi(j)) \quad & Z_i \in \mathbf{Z}^{\text{latent}}_n\cup\mathbf{Z}^{\text{prompt}}_n~\forall n \in [1, N] \land Z_j \notin \mathbf{Z}^{\text{prompt}} & \text{(6)} \\[8pt]
    \mathcal{B}(\pi(i)) & Z_i \in \mathbf{Z}^{\text{latent}} \land Z_j \in \mathbf{Z}^{\text{prompt}}_n~\forall n \in [1, N] & \text{(7)} \\[8pt]
    -\infty & \text{otherwise} & \quad \text{(8)}
    \end{array} \right.
\end{equation*}

In other words, we enforce strict binding between the tokens relative to the prompts (\textit{Case 1}) to prevent attribute leakage within the textual domain. 
Then, we keep the interaction of the global prompt and the background latents with all the latent and context tokens unbiased (\textit{Case 2}) to favor visual coherence in the unedited background regions, and do the same where the background latents and the context tokens attend to those relative to the global prompt (\textit{Case 3}). 
Furthermore, we apply a bias based on instance-aware smoothed masks to the interactions between prompt tokens and latent tokens, relative to each instance and to all latent tokens and context tokens (\textit{Case 6}). This ensures that the local, instance-specific prompt and the noisy image area enclosing each instance can interact with areas in the context image and the noisy image that are larger than the exact ones occupied by the instance, with a strength that is decreasing proportionally to the distance from the instance. In this way, the edited instance will blend better with its surroundings in the final image. 
The same biasing strategy is applied to the interaction between the latent tokens and the instance-specific prompts (\textit{Case 7}) to ensure coherent localization of the editing instruction.
As for the context tokens, we prevent those relative to the background from interacting with the instance prompts, and let the context tokens for each instance interact only with the corresponding instance-specific prompt (\textit{Case 4}). This choice is intended to better localize the instance-specific prompts within the reference image given as context. 
Finally, we keep the attention from the context tokens to the latent and context tokens unbiased (\textit{Case 5}) to favor layout, colors, and style preservation in the final image. 

\section{MICE-Bench}
\label{sec:mice_bench}

Existing benchmarks for multi-instance concurrent editing on natural images lack settings with a high number of concurrent edits. For example, the benchmark proposed by~\cite{yang2024object} contains only one or two edits per image, and~\cite{goirik_lomoe} is capped at seven edits, averaging only three. These benchmarks, therefore, do not provide significant avenues for verifying a model's ability to perform an arbitrarily high number of concurrent edits.

To bridge this gap, we introduce \textbf{MICE-Bench} (Multi-Instance Concurrent Editing Benchmark), which we specifically annotate for high-density editing. Notably, the benchmark features 260 samples with a significantly higher editing density than prior works, ranging from 5 to 40 concurrent edits per image, with an average of 8.5.
Formally, each sample can be defined as $(I^{src}, M, \mathbb T)$ where $I^{src}\in R^{H\times W\times 3}$ is the source image to be edited by the model, $M=\{m_n\}_{n=1}^N$ is the set of segmentation masks for each object in the image, and $\mathbb T$ is a set of prompts in the format "\textit{Replace the} \texttt{\{SRC\_OBJ\}} \textit{with} \texttt{\{TGT\_OBJ\}}", where \texttt{\{SRC\_OBJ\}} and \texttt{\{TGT\_OBJ\}} are respectively the name of the object to be edited in the source image and the name of the object that should replace it in the edited image. We provide some samples in the supplementary material.

\tit{Data Collection and Annotation}
We select 260 samples from the LVIS~\cite{gupta2019lvis} and COCO~\cite{lin2015coco} validation sets, focusing on having a diverse object count per sample ranging from 5 to 40 instances associated with segmentation masks and source object class names.
For each of them, we feed a multimodal LLM (Gemini 3 Flash) the source image, object class names, and normalized bounding boxes $[y_{min}, x_{min}, y_{max}, x_{max}]$ enclosing the original segmentation masks. The model is prompted to suggest creative yet contextually realistic replacements for every masked object. We manually verify these suggestions to ensure target feasibility within the scene (\eg~a "chipmunk" fits where a "mouse" is, whereas an "elephant" does not). 
We also use Gemini 3 Flash to generate an additional set of instructions explicitly containing textual information to identify each instance within the image (\eg~referring expressions, locations).
These are generated both as individual instructions and as a single global prompt.
Thanks to the multi-faceted annotation strategy (some samples are reported in the supplementary), MICE-Bench can be used to evaluate a wide range of approaches for multi-instance concurrent editing. 
\section{Experimental Analysis}
\label{sec:experiments}
\tit{Experiment setup} 
To evaluate our approach for multi-instance concurrent editing, we resort to our proposed MICE-Bench dataset (\cref{sec:mice_bench}), and to the recently-proposed LoMOE-Bench~\cite{goirik_lomoe} dataset, which contains 64 images associated with 2 to 7 editing instructions. 
To measure the performance, we adopt the following scores proposed in previous approaches tackling the same task~\cite{goirik_lomoe,zaccagnino2026shifting}:
\begin{itemize}[noitemsep, topsep=0pt,leftmargin=*]
    \item Target CLIP Score (\textbf{Tgt C}), \ie~the CLIP score~\cite{hessel2021clipscore} computed between the global prompt describing all the edits and the entire edited image~\cite{hessel2021clipscore}. This score captures the overall semantic adherence to the editing instructions. 
    \item Localized CLIP Score (\textbf{Loc C}), \ie~the CLIP score~\cite{hessel2021clipscore} computed between the target object description in the editing instruction and the bounding box areas around the edited regions. This score measures more precisely the capability of the model to perform each edit on each intended instance.
    \item Mean Absolute Error on the Background pixels (\textbf{MAE$_\text{B}$}), to estimate the background preservation capabilities of the model under investigation.
    \item Attempt Rate (\textbf{AR$_\text{\%}$}), \ie~the percentage of edits that the model has attempted to perform. Specifically, an edit is considered attempted when the pixel-wise difference between the original and edited areas relative to an instance exceeds a fixed threshold.
\end{itemize}

\tit{Implementation Details} 
We develop our training-free MICE method on the state-of-the-art editing model FLUX.2 [klein] 4B~\cite{flux-2-2025} and use it for the majority of our experiments, setting 4 inference steps. For MMDiT-based porting experiments, we choose the FLUX.2 [klein] 9B~\cite{flux-2-2025}, FLUX.1-Kontext-dev~\cite{labs2025flux_kontext} 12B, and FLUX.2-Dev~\cite{flux-2-2025} 32B. We use weights and reference code from HuggingFace~\cite{von-platen-etal-2022-diffusers}, setting 4 inference steps for FLUX.2 [klein] 9B and 28 steps for FLUX.1-Kontext-dev and FLUX.2-Dev. We use these models end-to-end, prompting them with textual prompts describing all the inputs (an analysis on the most suitable prompting strategy is reported in the supplementary material).
For smoothing the segmentation masks of the instances, we set kernel size $k$=3 and, due to differences in the edit types in the two benchmark datasets, temperature $t$=1 for LoMOE-Bench and $t$=8 for MICE-Bench. Please refer to the supplementary for further discussions. 
To compare MICE against alternative multi-instance editing approaches, we select representative methods for the categories described in~\Cref{sec:related}. In particular, we adapt the archetypal multi-branch MultiDiffusion~\cite{bar2023multidiffusion}, originally proposed for region-based generation, and apply it to FLUX.2 [klein] 4B~\cite{flux-2-2025} by using the official code as reference.\footnote{\url{https://github.com/omerbt/MultiDiffusion}} We run the multi-branch LoMOE~\cite{goirik_lomoe}, based on Stable Diffusion 2~\cite{rombach2022high} by using the official implementation\footnote{\url{https://github.com/goirik-chakrabarty/LoMOE}} and also port it on FLUX.2 [klein] 4B~\cite{flux-2-2025}. Moreover, we run the multi-branch LayerEdit~\cite{fu2025layeredit}, based on Stable Diffusion XL~\cite{podell2023sdxl} as released by the respective authors.\footnote{\url{https://github.com/fufy1024/LayerEdit}} We also implement the regularization-based multi-instance edit approach IDAttn~\cite{zaccagnino2026shifting}, based on FLUX.1-Kontext-dev~\cite{labs2025flux_kontext}, and port it on FLUX.2 [klein] 4B. 
Finally, we compare against the open-source Qwen-Image-Edit-2511~\cite{wu2025qwen_image} and the closed-source Gemini 3 Pro Image. 
We run all experiments on NVIDIA A100 40GB, with the exception of the commercial Gemini 3 Pro Image model, which we prompt by using the public Vertex API. 

\subsection{Results}
\tit{Effect of the Editing Area Definition}
In~\Cref{tab:ablation_smoothing}, we report an ablation on the effect of the signal used for instance localization. We compare Bounding Boxes (as used by~\cite{zaccagnino2026shifting} or in the supplementary prompting experiments), strict segmentation masks (as used by~\cite{goirik_lomoe,fu2025layeredit}), na\"{\i}vely-smoothed segmentation masks (using the Gaussian kernel), and masks obtained with our proposed instance-aware strategy. From the results, we can see how simply using masks instead of Bounding Boxes allows the model to better localize the edits while preserving the background. When introducing na\"{\i}ve smoothing, the model background preservation capabilities increase at the cost of reduce localized edits. This is mainly due to the overlap between instances and is therefore more detrimental on Mice-Bench. Adding instance-aware smoothing fixes this, and achieves the best performance, as the model is capable of localizing edits well while also harmonizing them in the background, without unwanted inter-instance leakages. 

\begin{table*}[t]
\centering
\caption{Ablation analysis of different editing areas definition strategies based on the segmentation masks (Seg. M.). Here, we apply MICE to FLUX.2 [klein] 4B.\vspace{-0.3cm}}
\label{tab:ablation_smoothing}
\resizebox{\linewidth}{!}{%
\begin{tabular}{l @{\hskip 1.5em} cccc @{\hskip 1.5em} cccc}
\toprule
& \multicolumn{4}{c}{\textbf{LoMOE}} & \multicolumn{4}{c}{\textbf{MICE-Bench}} \\
\cmidrule(r{1.5em}){2-5} \cmidrule{6-9}
& \textbf{Tgt C $\uparrow$} & \textbf{Loc C $\uparrow$} & \textbf{MAE$_\text{B}$ $\downarrow$} & \textbf{AR [\%] $\uparrow$} & \textbf{Tgt C $\uparrow$} & \textbf{Loc C $\uparrow$} & \textbf{MAE$_\text{B}$ $\downarrow$} & \textbf{AR [\%] $\uparrow$} \\
\midrule
Bounding Boxes &  25.94&  29.55&  9.38&  97.92&  25.00&  26.57&  10.66&  98.78\\
Original Seg. M. & 25.78& 29.59& 7.60& 96.35& 24.95& 26.63& 8.15& 95.36\\
Gaussian Seg. M. & 25.77& 29.55& 7.08& 96.35& 25.00& 26.19& 7.63& 87.30\\
Our Seg. M. & 25.77& 29.57& 7.07& 96.35& 25.15& 26.85& 7.84& 92.48\\
\bottomrule
\end{tabular}%
}\vspace{-0.3cm}
\end{table*}

\tit{Adaptation Strategies for Multi-Instance Concurrent Editing} 
In~\Cref{tab:adaptations}, we compare different training-free strategies proposed in the literature to obtain a multi-instance concurrent editing behavior from generative models that are trained for single-instance editing. Specifically, we implement the strategies proposed in~\cite{bar2023multidiffusion} (\textbf{MultiDiffusion}), in~\cite{goirik_lomoe} (\textbf{LoMOE}), and in~\cite{zaccagnino2026shifting} (\textbf{IDAttn}) on top of the same backbone (FLUX.2 [klein] 4B). The combination of multi-branch diffusion paths and gradient-based optimizations of LoMOE results in better performance on the simpler LoMOE-Bench dataset. 
However, MICE performs better on the more complex MICE-Bench in terms of localized edit quality. We argue that the reason is two fold. First, multi-branch approaches applied to few 
\begin{wrapfigure}[18]{r}{0.48\linewidth}
    \centering
    \vspace{-10pt}
    \includegraphics[width=\linewidth]{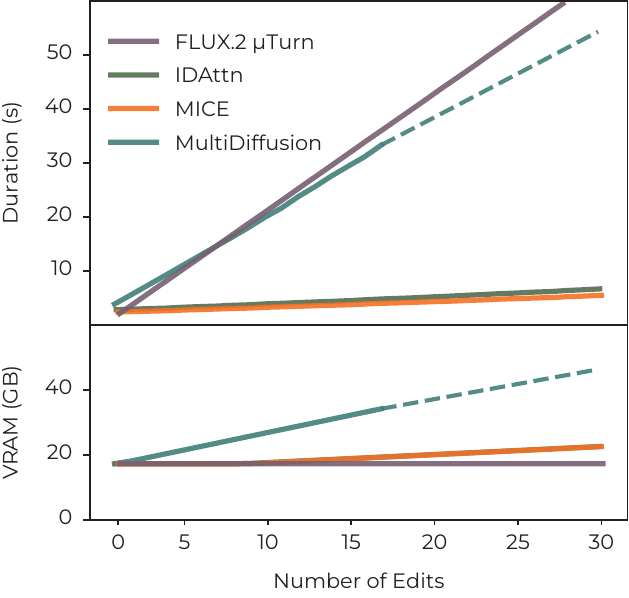}
    \caption{Inference time and VRAM comparison \wrt~the number of edits.}
    \label{fig:vram_time_scaling}
\end{wrapfigure}
sampling steps lack enough interaction points, especially with increasing instance number (and, therefore, number of branches to merge). Second, performing latent optimization for multiple instances creates numerical instabilities. This is also shown by the superior performance on MICE-Bench of MultiDiffusion, which has a pipeline similar to LoMOE but without gradient optimization. Regarding AR\% and MAE$_B$, we remark that they are not always correlated with better overall performance in the supplementary.
Moreover, we consider 1024$^2$ images and compute inference time and VRAM usage with increasing number of edits in~\Cref{fig:vram_time_scaling}, using a 40GB A100. As we can see, our method scales similarly to IDAttn. 
In contrast, MultiDiffusion scales linearly both in VRAM usage and inference time (we extrapolate the scaling behavior on $>$15 edits, as it does not fit in GPU memory). 
In summary, even with a few edits at practical resolutions, end-to-end regularized approaches are necessary when using current state-of-the-art editing models.  In the supplementary, we further elaborate on computational scaling behavior by presenting an asymptotic analysis of different approaches.  

\begin{table*}[t]
\centering
\caption{Comparison of different training-free strategies applied to FLUX.2 [klein] 4B to achieve multi-instance concurrent editing capabilities.}\vspace{-0.3cm}
\label{tab:adaptations}
\resizebox{\linewidth}{!}{%
\begin{tabular}{l @{\hskip 1.5em} cccc @{\hskip 1.5em} cccc}
\toprule
& \multicolumn{4}{c}{\textbf{LoMOE-Bench}} & \multicolumn{4}{c}{\textbf{MICE-Bench}} \\
\cmidrule(r{1.5em}){2-5} \cmidrule{6-9}
\textbf{Strategy} & \textbf{Tgt C $\uparrow$} & \textbf{Loc C $\uparrow$} & \textbf{MAE$_\text{B}$ $\downarrow$} & \textbf{AR [\%] $\uparrow$} & \textbf{Tgt C $\uparrow$} & \textbf{Loc C $\uparrow$} & \textbf{MAE$_\text{B}$ $\downarrow$} & \textbf{AR [\%] $\uparrow$} \\
\midrule
MultiDiffusion & 26.04& 30.12& 6.39& 95.31& 24.79& 26.26& 6.21& 85.25\\
LoMOE & 26.23& 30.15& 8.32& 98.44& 24.85& 26.04& 8.62& 91.46\\
IDAttn & 25.29& 29.12& 7.04& 92.19& 24.64& 25.54 & 6.42& 77.05\\
MICE & 25.77& 29.57& 7.07& 96.35& 25.15 & 26.85 & 7.84& 92.48\\
\bottomrule
\end{tabular}%
}
\end{table*}

\tit{Backbone Adaptability} 
Our approach can be easily integrated in different MMDiT-based editing models. Note that the $k$ and $\tau$ hyperparameters can always be adjusted depending on the type of edit to perform, but our approach does not rely on architecture-related hyperparameters such as the layers where to apply our attention biasing strategy. To evaluate the portability of MICE, we apply it to different state-of-the-art image editing backbones and report the results of this analysis in~\Cref{tab:backbones}. 
First, we consider FLUX.2 [klein] 9B, whose inputs and outputs are passed through the same blocks as in the 4B version, but whose MMDiT is larger. Moreover, we consider the \textbf{FLUX.2-Dev} variant, which has a different preprocessing pipeline compared to the [klein] variants (different text and image encoder) and a larger MMDiT. Finally, we consider \textbf{FLUX.1 Kontext}, which has yet another preprocessing pipeline compared to the others. As we can see, our MICE method shows consistent improvements in localized edit quality across all baselines on the two benchmarks, indicated by Loc C. In the remaining experiments, we use the FLUX.2 [klein] 9B porting (dubbed MICE-9B).

\begin{table*}[t]
\centering
\caption{Performance of various MMDiT-based generative editing backbones with the integration of our approach (+ MICE).}\vspace{-0.3cm}
\label{tab:backbones}
\resizebox{\linewidth}{!}{%
\begin{tabular}{l cccc cccc}
\toprule
& \multicolumn{4}{c}{\textbf{LoMOE-Bench}} & \multicolumn{4}{c}{\textbf{MICE-Bench}} \\
\cmidrule(r{1.5em}){2-5} \cmidrule{6-9}
& \textbf{Tgt C $\uparrow$} & \textbf{Loc C $\uparrow$} & \textbf{MAE$_\text{B}$ $\downarrow$} & \textbf{AR [\%] $\uparrow$} & \textbf{Tgt C $\uparrow$} & \textbf{Loc C $\uparrow$} & \textbf{MAE$_\text{B}$ $\downarrow$} & \textbf{AR [\%] $\uparrow$} \\
\midrule
FLUX.1 Kontext & 25.77 & 28.26 & 9.46 & 93.23 & 25.65 & 25.43 & 16.75 & 97.95 \\
~~+ MICE & 25.42 & 29.13 & 7.49 & 94.27 & 24.54 & 25.54 & 6.61 & 86.43 \\
\midrule
FLUX.2 [klein] 4B & 25.94 & 28.68 & 8.76 & 97.40 & 25.37 & 25.97 & 26.08 & 99.61 \\
~~+ MICE & 25.77 & 29.57 & 7.07 & 96.35 & 25.15 & 26.85 & 7.84 & 92.48 \\
\midrule
FLUX.2 [klein] 9B & 26.57 & 29.45 & 8.63 & 99.48 & 25.66 & 26.34 & 27.12 & 100.00 \\
~~+ MICE & 26.49 & 30.30 & 8.88 & 98.44 & 25.24 & 27.79 & 13.89 & 98.97 \\
\midrule
FLUX.2-Dev & 27.03 & 29.00 & 10.48 & 97.92 & 25.93 & 26.25 & 19.24 & 99.61 \\
~~+ MICE & 26.44 & 30.28 & 11.50 & 99.48 & 25.09 & 26.91 & 22.54 & 98.97 \\
\bottomrule
\end{tabular}%
}\vspace{-0.3cm}
\end{table*}

\tit{Comparison with the State-of-the-Art} 
\Cref{tab:leaderboard} contains the performance of MICE-9B when compared to end-to-end single pass image editing models, including our baseline model (Flux.2 [klein] 9B), state-of-the-art open (\textbf{Qwen-Image-Edit}~\cite{wu2025qwen_image}) and closed (\textbf{Gemini 3 Pro Image}~\cite{comanici2025gemini}) source models provided with textual prompts only. 
Moreover, we consider recent strategies specifically proposed for multi-instance concurrent editing, on the original backbone proposed by the respective authors. These are \textbf{LoMOE}~\cite{goirik_lomoe} and \textbf{LayerEdit}~\cite{fu2025layeredit} (which rely on the segmentation masks of the instances to edit, as MICE does), and \textbf{IDAttn}~\cite{zaccagnino2026shifting} (which takes as input the bounding boxes surrounding the instances). In this comparison, we also include the results obtained by our approach when given instance segmentation masks obtained by prompting the instance segmentation model SAM3~\cite{carion2025sam} instead of the ground truth ones from the datasets. As we can see, the superior performance of our MICE method in Localized CLIP score indicates that the model successfully performs localized edits, even beating Gemini 3 Pro, which showed better performance than the FLUX.2 [klein] 9B end-to-end baseline. Moreover, our method maintains superior performance also when fed with the segmentation masks from SAM3, showing its usability in real-world scenarios where a user can simply provide points as visual cues instead of precise segmentation masks. Further analysis is reported in the supplementary material.

\begin{table*}[t]
\centering
\caption{Comparison with State-of-the-Art Editing Models.}\vspace{-0.3cm}
\label{tab:leaderboard}
\resizebox{\linewidth}{!}{%
\begin{tabular}{l cccc cccc}
\toprule
& \multicolumn{4}{c}{\textbf{LoMOE}} & \multicolumn{4}{c}{\textbf{MICE-Bench}} \\
\cmidrule(r{1.5em}){2-5} \cmidrule{6-9}
\textbf{Model} & \textbf{Tgt C $\uparrow$} & \textbf{Loc C $\uparrow$} & \textbf{MAE$_\text{B}$ $\downarrow$} & \textbf{AR [\%] $\uparrow$} & \textbf{Tgt C $\uparrow$} & \textbf{Loc C $\uparrow$} & \textbf{MAE$_\text{B}$ $\downarrow$} & \textbf{AR [\%] $\uparrow$} \\
\midrule
FLUX.2 [klein] 9B & 26.57 & 29.45 & 8.63 & 99.48 & 25.66 & 26.34 & 27.12 & 100.00 \\
Gemini 3 Pro Image & 26.82 & 29.67 & 7.24 & 97.37 & 25.64 & 27.09 & 13.08 & 98.97 \\
IDAttn & 25.67 & 29.26 & 6.41 & 98.44 & 24.68 & 25.54 & 5.06 & 73.01 \\
LayerEdit & 25.61 & 29.07 & 7.43 & 97.92 & 23.77 & 24.38 & 9.32 & 99.51 \\
LoMOE & 26.00 & 29.40 & 6.66 & 92.19 & 24.99 & 25.46 & 7.76 & 99.07 \\
Qwen-Image-Edit & 26.34 & 28.85 & 7.13 & 88.02 & 25.26 & 25.33 & 26.47 & 98.83 \\
\midrule
MICE & 26.49 & 30.30 & 8.88 & 98.44 & 25.24 & 27.79 & 13.89 & 98.97 \\
\hspace{1em} + SAM3 & 26.22 & 30.41 & 8.94 & 96.88 & 25.26 & 27.73 & 13.36 & 98.54 \\
\bottomrule
\end{tabular}%
}\vspace{-0.4cm}
\end{table*}

\begin{figure}[t]
    \centering
    \includegraphics[width=\linewidth]{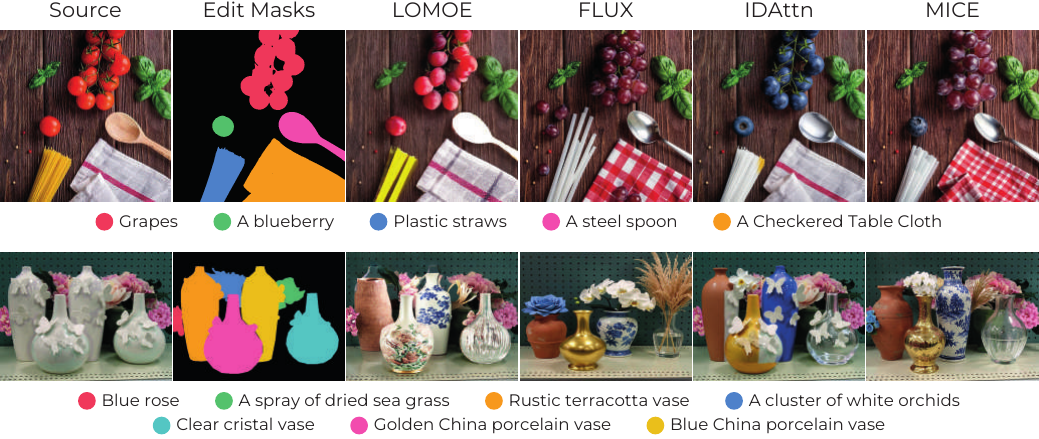}\vspace{-0.2cm}
    \caption{Qualitative Results on LoMOE-Bench and MICE-Bench.}\vspace{-0.5cm}
    \label{fig:qualitatives}
\end{figure}

\tit{Qualitative Analysis}
We report qualitative results in~\Cref{fig:qualitatives}. In both samples, LoMOE shows overall low generation quality, even failing convincingly generate any of the instances in the LoMOE-Bench sample. FLUX.2 [klein] in end-to-end mode fares much better, but fails to preserve the overall composition and misplaces some objects. 
IDAttn, on the other hand, misses some edits, while retaining the overall composition and object placement. MICE shines in background and layout preservation as well as in prompt following in both samples.
We perform an LLM-as-judge evaluation using Gemini 3.1 Pro with high thinking budget, with 30 samples for each dataset, passing to the LLM judge the source image, the edit locations, the edit prompts and the result from two models, asking it to choose a winner among the two in terms of prompt following, background preservation and overall visual quality. We include in this evaluation our model, the FLUX.2 [klein] baseline it is based on, and open-source competitors 
\begin{wraptable}[8]{r}{0.48\linewidth}
    \centering
    \vspace{-25pt} 
    \caption{LLM-as-judge Results.}
    \label{tab:merged_eval_vertical}
    \resizebox{\linewidth}{!}{%
        \begin{tabular}{l c c}
        \toprule
         & \textbf{LoMOE-Bench} & \textbf{MICE-Bench} \\
        \midrule
        FLUX.2 [klein] 9B  & 1299.78 & 1172.16 \\
        IDAttn          & 1223.57 & 1412.73 \\
        LayerEdit       & 887.98  & 858.92  \\
        LoMOE           & 1181.02 & 995.05  \\
        Qwen-Image-Edit & 1265.12 & 1192.69 \\
        MICE   & 1342.53 & 1568.46 \\
        \bottomrule
        \end{tabular}%
    }
\end{wraptable}
Qwen-Image-Edit-2511, and LoMOE, LayerEdit and IDAttn, asking the LLM 900 total binary choice questions. We compute ELO scores from these two-way match-ups.
Note that~\cite{haraguchi2024can, zaccagnino2026shifting} showed good correlation between user and LLM responses to questions about evaluating image visual features and ranking models based on image editing ability. Moreover, to substantiate this correlation in our case, we also conduct a user study, for which we report results in the supplementary. 
We report the results of the LLM-as-judge evaluation in~\Cref{tab:merged_eval_vertical}. On both datasets, MICE ends up being the best-performing model. The second-best on LoMOE is FLUX.2 [klein], followed by Qwen-Image-Edit and IDAttn, with LoMOE and LayerEdit performing significantly worse in this evaluation. When switching to MICE-Bench, the denser annotation favors the specialized IDAttn model, whereas the end-to-end competitors FLUX.2 [klein] and Qwen-Image-Edit fall much further behind. As highlighted by the authors of~\cite{zaccagnino2026shifting}, the bad generation quality of LayerEdit and LoMOE especially fails to be highlighted by CLIP score, but is made evident by human and LLM-based evaluation.

\begin{figure}[h]
    \centering
    \includegraphics[width=\linewidth]{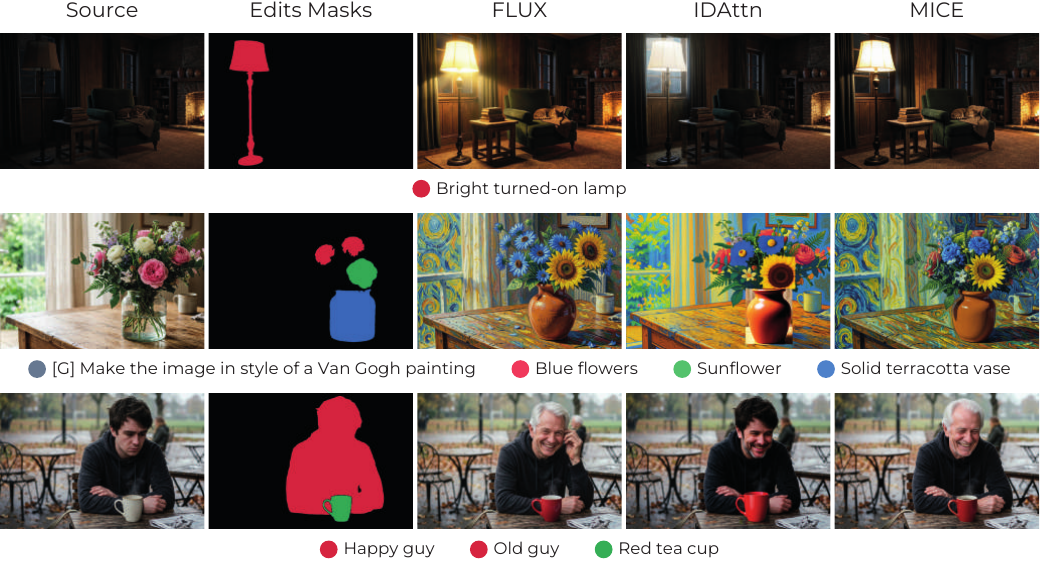}\vspace{-0.2cm}
    \caption{Exemplar output of FLUX.2 [klein], IDAttn, and MICE on the edge cases}\vspace{-0.5cm}
    \label{fig:edge_cases_discussion_main}
\end{figure}

\tit{Edge Cases} Additionally, in~\Cref{fig:edge_cases_discussion_main} we show a qualitative analysis of our proposed method in some edge cases. In particular, we show how MICE does not hinder the base FLUX.2 [klein] model capabilities of performing global edits and harmonizing the resulting image, either when the global harmonization is necessary due to the localized prompt (first image, where illumination changes are necessary after the local edit) or when the edit instructions require combining global style changes with local edits (second image). Finally, in the third image, we show that our method can handle multiple edits affecting the same region. Arguably, this is due to MICE's smoothly-disentangled attention masking strategy, especially considering the output of IDAttn.

\section{Conclusion}
In this work, we tackle the problem of multi-instance concurrent image editing by proposing MICE, a training-free and architecture-agnostic method that leverages the joint attention mechanism of MMDiTs by redefining the additive bias applied to attention logits. Specifically, MICE alters token interactions according to instance-specific regions to promote attribute binding within instances while maintaining sufficient interactions across the image to preserve its global coherence.
Moreover, we have introduced MICE-Bench, a new dataset designed to assess performance under more demanding conditions, \ie~a higher number of concurrent edits per sample compared to existing benchmarks for multi-instance image editing. Extensive evaluation across diverse datasets and backbones demonstrates that our method consistently improves instruction adherence, edit localization, and background preservation compared with recent baselines and competitors. These results suggest that attention biasing provides an effective mechanism to scale multi-instance editing to complex scenarios.

\section*{Acknowledgment}
This paper is based upon work supported by the ``AI for Digital Humanities'' project funded by ``Fondazione di Modena'', the GCP Credit Award, the Google Cloud Research Credits program with the award GCP19980904, and the FARD2025 project (CUP E93C25000370005).
We acknowledge EuroHPC Joint Undertaking and ISCRA for awarding us access to LUMI at CSC, Finland, LEONARDO at CINECA, Italy, and MareNostrum5 at BSC, Spain. 

%
%
\bibliographystyle{splncs04}
\bibliography{main}

@String(CVPR= {IEEE Conf. Comput. Vis. Pattern Recog.})

@String(ICCV= {Int. Conf. Comput. Vis.})

@String(ECCV= {Eur. Conf. Comput. Vis.})

@String(NIPS= {Adv. Neural Inform. Process. Syst.})

@String(ACMMM= {ACM Int. Conf. Multimedia})

@String(ICASSP=	{ICASSP})

@String(ICLR = {Int. Conf. Learn. Represent.})

@String(AAAI = {AAAI})

@String(JMLR = {J. Mach. Learn. Res.})

@String(WACV = {Winter Conference on Applications of Computer Vision})

@String(CVPR  = {CVPR})

@String(ICCV  = {ICCV})

@String(ECCV  = {ECCV})

@String(NIPS  = {NeurIPS})

@String(ACMMM = {ACM MM})

@String(ICLR  = {ICLR})

@String(JMLR = {JMLR})

@String(WACV = {WACV})

@String(ICML = {ICML})

@String(EMNLP = {EMNLP})

@article{raffel2020exploring,
  title={{Exploring the Limits of Transfer Learning with a Unified Text-to-Text Transformer}},
  author={Raffel, Colin and Shazeer, Noam and Roberts, Adam and Lee, Katherine and Narang, Sharan and Matena, Michael and Zhou, Yanqi and Li, Wei and Liu, Peter J},
  journal=JMLR,
  pages={1--67},
  year={2020}
}

@inproceedings{bar2023multidiffusion,
  title={{MultiDiffusion: Fusing Diffusion Paths for Controlled Image Generation}},
  author={Bar-Tal, Omer and Yariv, Lior and Lipman, Yaron and Dekel, Tali},
  booktitle=icml,
  year={2023},
}

@inproceedings{dahary2024beyourself,
  title={{Be Yourself: Bounded Attention for Multi-Subject Text-to-Image Generation}},
  author={Dahary, Omer and Patashnik, Or and Aberman, Kfir and Cohen-Or, Daniel},
  booktitle=ECCV,
  pages={432--448},
  year={2024},
  organization={Springer}
}

@inproceedings{haraguchi2024can,
  title={{Can GPTs evaluate graphic design based on design principles?}},
  author={Haraguchi, Daichi and Inoue, Naoto and Shimoda, Wataru and Mitani, Hayato and Uchida, Seiichi and Yamaguchi, Kota},
  booktitle={SIGGRAPH},
  pages={1--4},
  year={2024}
}

@article{liu2025step1x,
  title={{Step1X-Edit: A Practical Framework for General Image Editing}},
  author={Liu, Shiyu and Han, Yucheng and Xing, Peng and Yin, Fukun and Wang, Rui and Cheng, Wei and Liao, Jiaqi and Wang, Yingming and Fu, Honghao and Han, Chunrui and others},
  journal={arXiv preprint arXiv:2504.17761},
  year={2025}
}

@inproceedings{tan2025ominicontrol,
  title={{OminiControl: Minimal and Universal Control for Diffusion Transformer}},
  author={Tan, Zhenxiong and Liu, Songhua and Yang, Xingyi and Xue, Qiaochu and Wang, Xinchao},
  booktitle=ICCV,
  pages={14940--14950},
  year={2025}
}

@article{wu2025qwen_image,
  title={{Qwen-Image Technical Report}},
  author={Wu, Chenfei and Li, Jiahao and Zhou, Jingren and Lin, Junyang and Gao, Kaiyuan and Yan, Kun and Yin, Sheng-ming and Bai, Shuai and Xu, Xiao and Chen, Yilei and others},
  journal={arXiv preprint arXiv:2508.02324},
  year={2025}
}

@article{wang2025seededit,
  title={{SeedEdit 3.0: Fast and High-Quality Generative Image Editing}},
  author={Wang, Peng and Shi, Yichun and Lian, Xiaochen and Zhai, Zhonghua and Xia, Xin and Xiao, Xuefeng and Huang, Weilin and Yang, Jianchao},
  journal={arXiv preprint arXiv:2506.05083},
  year={2025}
}

@inproceedings{simsar2025lime,
  title={{LIME: Localized Image Editing via Attention Regularization in Diffusion Models}},
  author={Simsar, Enis and Tonioni, Alessio and Xian, Yongqin and Hofmann, Thomas and Tombari, Federico},
  booktitle=WACV,
  pages={222--231},
  year={2025},
  organization={IEEE}
}

@inproceedings{zhang2025eligen,
  title={{EliGen: Entity-Level Controlled Image Generation with Regional Attention}},
  author={Zhang, Hong and Duan, Zhongjie and Wang, Xingjun and Chen, Yingda and Zhang, Yu},
  booktitle={ACM-MMAsia},
  year={2025}
}

@article{comanici2025gemini,
  title={{Gemini 2.5: Pushing the Frontier with Advanced Reasoning, Multimodality, Long Context, and Next Generation Agentic Capabilities}},
  author={Comanici, Gheorghe and Bieber, Eric and Schaekermann, Mike and Pasupat, Ice and Sachdeva, Noveen and Dhillon, Inderjit and Blistein, Marcel and Ram, Ori and Zhang, Dan and Rosen, Evan and others},
  journal={arXiv preprint arXiv:2507.06261},
  year={2025}
}

@article{labs2025flux_kontext,
  title={{FLUX. 1 Kontext: Flow Matching for In-Context Image Generation and Editing in Latent Space}},
  author={Labs, Black Forest and Batifol, Stephen and Blattmann, Andreas and Boesel, Frederic and Consul, Saksham and Diagne, Cyril and Dockhorn, Tim and English, Jack and English, Zion and Esser, Patrick and others},
  journal={arXiv preprint arXiv:2506.15742},
  year={2025}
}

@article{zhu2025mde,
  title={{MDE-Edit: Masked Dual-Editing for Multi-Object Image Editing via Diffusion Models}},
  author={Zhu, Hongyang and Liu, Haipeng and Fu, Bo and Wang, Yang},
  journal={arXiv preprint arXiv:2505.05101},
  year={2025}
}

@inproceedings{matsuda2024multi,
  title={{Multi-object editing in personalized text-to-image diffusion model via segmentation guidance}},
  author={Matsuda, Haruka and Togo, Ren and Maeda, Keisuke and Ogawa, Takahiro and Haseyama, Miki},
  booktitle={ICASSP},
  pages={8140--8144},
  year={2024},
  organization={IEEE}
}

@inproceedings{goirik_lomoe,
author = {Chakrabarty, Goirik and Chandrasekar, Aditya and Hebbalaguppe, Ramya and AP, Prathosh},
title = {{LoMOE: Localized Multi-Object Editing via Multi-Diffusion}},
year = {2024},
booktitle = ACMMM,
pages = {3342–3351}
}

@inproceedings{yang2024object,
  title={{Object-Aware Inversion and Reassembly for Image Editing}},
  author={Yang, Zhen and Ding, Ganggui and Wang, Wen and Chen, Hao and Zhuang, Bohan and Shen, Chunhua},
  booktitle=ICLR,
  year={2024}
}

@inproceedings{li2025moedit,
  title={{MoEdit: On Learning Quantity Perception for Multi-object Image Editing}},
  author={Li, Yanfeng and Chan, Kahou and Sun, Yue and Lam, Chantong and Tong, Tong and Yu, Zitong and Fu, Keren and Liu, Xiaohong and Tan, Tao},
  booktitle=CVPR,
  pages={2683--2693},
  year={2025}
}

@article{tong2024improving,
  title={Improving and generalizing flow-based generative models with minibatch optimal transport},
  author={Tong, Alexander and Fatras, Kilian and Malkin, Nikolay and Huguet, Guillaume and Zhang, Yanlei and Rector-Brooks, Jarrid and Wolf, Guy and Bengio, Yoshua},
  journal={TMLR},
  pages={1--34},
  year={2024}
}

@article{zheng2023guided,
  title={Guided flows for generative modeling and decision making},
  author={Zheng, Qinqing and Le, Matt and Shaul, Neta and Lipman, Yaron and Grover, Aditya and Chen, Ricky TQ},
  journal={arXiv preprint arXiv:2311.13443},
  year={2023}
}

@article{zhou2025dreamrenderer,
  title={{DreamRenderer: Taming multi-instance attribute control in large-scale text-to-image models}},
  author={Zhou, Dewei and Li, Mingwei and Yang, Zongxin and Yang, Yi},
  journal=ICCV,
  year={2025}
}

@article{chen2024region,
  title={Region-aware text-to-image generation via hard binding and soft refinement},
  author={Chen, Zhennan and Li, Yajie and Wang, Haofan and Chen, Zhibo and Jiang, Zhengkai and Li, Jun and Wang, Qian and Yang, Jian and Tai, Ying},
  journal=ICCV,
  year={2025}
}

@inproceedings{eijkelboomcontrolled,
  title={{Controlled Generation with Equivariant Variational Flow Matching}},
  author={Eijkelboom, Floor and Zimmermann, Heiko and Vadgama, Sharvaree and Bekkers, Erik J and Welling, Max and Naesseth, Christian A and van de Meent, Jan-Willem},
  booktitle=ICML,
  year={2025}
}

@article{he2024calmflow,
  title={{CaLMFlow: Volterra Flow Matching using Causal Language Models}},
  author={He, Sizhuang and Levine, Daniel and Vrkic, Ivan and Bressana, Marco Francesco and Zhang, David and Rizvi, Syed Asad and Zhang, Yangtian and Zappala, Emanuele and van Dijk, David},
  journal={arXiv preprint arXiv:2410.05292},
  year={2024}
}

@inproceedings{var2024,
  title={{Visual Autoregressive Modeling: Scalable Image Generation via Next-Scale Prediction}},
  author={Tian, Keyu and Jiang, Yi and Yuan, Zehuan and Jia, Bohan and Chen, Boxun},
  booktitle=NIPS,
  year={2024}
}

@inproceedings{infinity2024,
  title={{Infinity: Scaling Bitwise AutoRegressive Modeling for High-Resolution Image Synthesis}},
  author={Han, Jian and Liu, Jinlai and Jiang, Yi and Yan, Bin and Zhang, Yuqi and Yuan, Zehuan and Peng, Bingyue and Liu, Xiaobing},
  booktitle=CVPR,
  year={2025}
}

@article{controlvar2024,
  title={{ControlVAR: Exploring Controllable Visual Autoregressive Modeling}},
  author={Li, Xiang and Qiu, Kai and Chen, Hao and Kuen, Jason and Lin, Zhe and Singh, Rita and Raj, Bhiksha},
  journal={arXiv preprint arXiv:2406.09750},
  year={2024}
}

@article{contextar2025,
  title={{Context-Aware Autoregressive Models for Multi-Conditional Image Generation}},
  author={Chen, Yixiao and Ma, Zhiyuan and Jia, Guoli and Jiang, Che and Li, Jianjun and Zhou, Bowen},
  journal={arXiv preprint arXiv:2505.12274},
  year={2025}
}

@article{team2025nextstep,
  title={Nextstep-1: Toward autoregressive image generation with continuous tokens at scale},
  author={Team, NextStep and Han, Chunrui and Li, Guopeng and Wu, Jingwei and Sun, Quan and Cai, Yan and Peng, Yuang and Ge, Zheng and Zhou, Deyu and Tang, Haomiao and others},
  journal={arXiv preprint arXiv:2508.10711},
  year={2025}
}

@inproceedings{paralleledits2024,
  title={{ParallelEdits: Efficient Multi-Aspect Text-Driven Image Editing with Attention Grouping}},
  author={Huang, Mingzhen and Cai, Jialing and Jia, Shan and Lokhande, Vishnu Suresh and Lyu, Siwei},
  booktitle=nips,
  year={2024}
}

@inproceedings{esser2024scaling,
  title={{Scaling Rectified Flow Transformers for High-Resolution Image Synthesis}},
  author={Esser, Patrick and Kulal, Sumith and Blattmann, Andreas and Entezari, Rahim and M{\"u}ller, Jonas and Saini, Harry and Levi, Yam and Lorenz, Dominik and Sauer, Axel and Boesel, Frederic and others},
  booktitle=ICML,
  pages={12606--12633},
  year={2024},
  organization={PMLR}
}

@inproceedings{lipman2023flow,
  title={{Flow Matching for Generative Modeling}},
  author={Lipman, Yaron and Chen, Ricky TQ and Ben-Hamu, Heli and Nickel, Maximilian and Le, Matt},
  booktitle=ICLR,
  year={2023}
}

@inproceedings{rombach2022high,
  title={{High-resolution Image Synthesis with Latent Diffusion Models}},
  author={Rombach, Robin and Blattmann, Andreas and Lorenz, Dominik and Esser, Patrick and Ommer, Bj{\"o}rn},
  booktitle=cvpr,
  year={2022}
}

@inproceedings{podell2023sdxl,
title={{{SDXL}: Improving Latent Diffusion Models for High-Resolution Image Synthesis}},
author={Dustin Podell and Zion English and Kyle Lacey and Andreas Blattmann and Tim Dockhorn and Jonas M{\"u}ller and Joe Penna and Robin Rombach},
booktitle=iclr,
year={2024}
}

@misc{von-platen-etal-2022-diffusers,
  author = {Patrick von Platen and Suraj Patil and Anton Lozhkov and Pedro Cuenca and Nathan Lambert and Kashif Rasul and Mishig Davaadorj and Dhruv Nair and Sayak Paul and William Berman and Yiyi Xu and Steven Liu and Thomas Wolf},
  title = {Diffusers: State-of-the-art diffusion models},
  year = {2022},
  publisher = {GitHub},
  journal = {GitHub repository},
  howpublished = {\url{https://github.com/huggingface/diffusers}}
}

@misc{flux-2-2025,
    author={Black Forest Labs},
    title={{FLUX.2: Frontier Visual Intelligence}},
    year={2025},
    howpublished={\url{https://bfl.ai/blog/flux-2}},
}

@inproceedings{mun2025ale,
  title={Addressing Text Embedding Leakage in Diffusion-based Image Editing},
  author={Sunung Mun and Jinhwan Nam and Sunghyun Cho and Jungseul Ok},
  booktitle=iccv,
  year={2025}
}

@article{fu2025layeredit,
  title={{LayerEdit: Disentangled Multi-Object Editing via Conflict-Aware Multi-Layer Learning}},
  author={Fu, Fengyi and Huang, Mengqi and Zhang, Lei and Mao, Zhendong},
  journal=AAAI,
  year={2026}
}

@inproceedings{hessel2021clipscore,
  title={{CLIPScore:} A Reference-free Evaluation Metric for Image Captioning},
  author={Hessel, Jack and Holtzman, Ari and Forbes, Maxwell and Bras, Ronan Le and Choi, Yejin},
  booktitle=emnlp,
  year={2021}
}

@inproceedings{zaccagnino2026shifting,
  title={{Shifting the Breaking Point of Flow Matching for Multi-Instance Editing}},
  author={Zaccagnino, Carmine and Quattrini, Fabio and Simsar, Enis and Gazulla, Marta Tintor{\'e} and Cucchiara, Rita and Tonioni, Alessio and Cascianelli, Silvia},
  booktitle=ICML,
  year={2026}
}

@inproceedings{gupta2019lvis,
  title={LVIS: A Dataset for Large Vocabulary Instance Segmentation},
  author={Gupta, Agrim and Dollar, Piotr and Girshick, Ross},
  booktitle=cvpr,
  year={2019}
}

@inproceedings{wu2024general,
  title={General object foundation model for images and videos at scale},
  author={Wu, Junfeng and Jiang, Yi and Liu, Qihao and Yuan, Zehuan and Bai, Xiang and Bai, Song},
  booktitle=cvpr,
  year={2024}
}

@inproceedings{kirillov2023segment,
  title={Segment anything},
  author={Kirillov, Alexander and Mintun, Eric and Ravi, Nikhila and Mao, Hanzi and Rolland, Chloe and Gustafson, Laura and Xiao, Tete and Whitehead, Spencer and Berg, Alexander C and Lo, Wan-Yen and others},
  booktitle=iccv,
  year={2023}
}

@inproceedings{carion2025sam,
  title={{Sam 3: Segment anything with concepts}},
  author={Carion, Nicolas and Gustafson, Laura and Hu, Yuan-Ting and Debnath, Shoubhik and Hu, Ronghang and Suris, Didac and Ryali, Chaitanya and Alwala, Kalyan Vasudev and Khedr, Haitham and Huang, Andrew and others},
  booktitle=iclr,
  year={2026}
}

@misc{lin2015coco,
      title={Microsoft COCO: Common Objects in Context}, 
      author={Tsung-Yi Lin and Michael Maire and Serge Belongie and Lubomir Bourdev and Ross Girshick and James Hays and Pietro Perona and Deva Ramanan and C. Lawrence Zitnick and Piotr Dollár},
      year={2015},
      eprint={1405.0312},
      archivePrefix={arXiv},
      primaryClass={cs.CV},
      url={https://arxiv.org/abs/1405.0312}, 
}

@inproceedings{ye2025imgedit,
  title={ImgEdit: A Unified Image Editing Dataset and Benchmark},
  author={Ye, Yang and He, Xianyi and Li, Zongjian and Lin, Bin and Yuan, Shenghai and Yan, Zhiyuan and Hou, Bohan and Yuan, Li},
  booktitle=nips,
  year={2025}
}

@misc{qu2025replan,
      title={RePlan: Reasoning-guided Region Planning for Complex Instruction-based Image Editing}, 
      author={Tianyuan Qu and Lei Ke and Xiaohang Zhan and Longxiang Tang and Yuqi Liu and Bohao Peng and Bei Yu and Dong Yu and Jiaya Jia},
      year={2025},
      eprint={2512.16864},
      archivePrefix={arXiv},
      primaryClass={cs.CV},
      url={https://arxiv.org/abs/2512.16864}, 
}

@inproceedings{zhou2025multi,
  title={Multi-turn consistent image editing},
  author={Zhou, Zijun and Deng, Yingying and He, Xiangyu and Dong, Weiming and Tang, Fan},
  booktitle=iccv,
  year={2025}
}

@misc{yang2025qwen3,
      title={Qwen3 Technical Report}, 
      author={An Yang and Anfeng Li and Baosong Yang and Beichen Zhang and Binyuan Hui and Bo Zheng and Bowen Yu and Chang Gao and Chengen Huang and Chenxu Lv and Chujie Zheng and Dayiheng Liu and Fan Zhou and Fei Huang and Feng Hu and Hao Ge and Haoran Wei and Huan Lin and Jialong Tang and Jian Yang and Jianhong Tu and Jianwei Zhang and Jianxin Yang and Jiaxi Yang and Jing Zhou and Jingren Zhou and Junyang Lin and Kai Dang and Keqin Bao and Kexin Yang and Le Yu and Lianghao Deng and Mei Li and Mingfeng Xue and Mingze Li and Pei Zhang and Peng Wang and Qin Zhu and Rui Men and Ruize Gao and Shixuan Liu and Shuang Luo and Tianhao Li and Tianyi Tang and Wenbiao Yin and Xingzhang Ren and Xinyu Wang and Xinyu Zhang and Xuancheng Ren and Yang Fan and Yang Su and Yichang Zhang and Yinger Zhang and Yu Wan and Yuqiong Liu and Zekun Wang and Zeyu Cui and Zhenru Zhang and Zhipeng Zhou and Zihan Qiu},
      year={2025},
      eprint={2505.09388},
      archivePrefix={arXiv},
      primaryClass={cs.CL},
      url={https://arxiv.org/abs/2505.09388}, 
}

@Article{pau2024mathematical,
AUTHOR = {Pau, Danilo Pietro and Aymone, Fabrizio Maria},
TITLE = {Mathematical Formulation of Learning and Its Computational Complexity for Transformers’ Layers},
JOURNAL = {Eng},
VOLUME = {5},
YEAR = {2024},
NUMBER = {1},
PAGES = {34--50},
URL = {https://www.mdpi.com/2673-4117/5/1/3},
ISSN = {2673-4117},
DOI = {10.3390/eng5010003}
}
\end{document}


\appendix
\title{Supplementary Material for:\texorpdfstring{\\}{ }Editing Everything Everywhere All at Once }

\author{Fabio Quattrini\thanks{Equal contribution, order determined by a coin flip.}\inst{1}\orcidlink{0009-0004-3244-6186} \and
Carmine Zaccagnino$^{\star}$\inst{1}\orcidlink{0009-0004-4953-6348} \and
Enis Simsar\inst{2}\orcidlink{0000-0002-6662-3249}  \and
Marta Tintoré Gazulla\inst{3} \and
Rita Cucchiara\inst{1}\orcidlink{0000-0002-2239-283X} \and
Alessio Tonioni\inst{3}\orcidlink{0000-0003-3358-9686} \and
Silvia Cascianelli\inst{1}\orcidlink{0000-0001-7885-6050}}

\authorrunning{F.~Quattrini, C. Zaccagnino et al.}

\institute{University of Modena and Reggio Emilia \and ETH Zurich \and Google}

\begingroup
\renewcommand{\addcontentsline}[3]{}
\renewcommand{\addtocontents}[2]{}
\maketitle
\endgroup

\begingroup
\let\clearpage\relax
\hypersetup{linkcolor=black}
\vspace{2.5em} 
\setcounter{tocdepth}{2}
\makeatletter
\@starttoc{toc}
\makeatother
\vspace{1.5em} 
\endgroup

\setlength{\belowdisplayskip}{3pt} \setlength{\belowdisplayshortskip}{3pt}
\setlength{\abovedisplayskip}{3pt} \setlength{\abovedisplayshortskip}{3pt}

\section{Effect of $k$ and $\tau$}

\begin{figure}
    \centering
    \includegraphics[width=.9\linewidth]{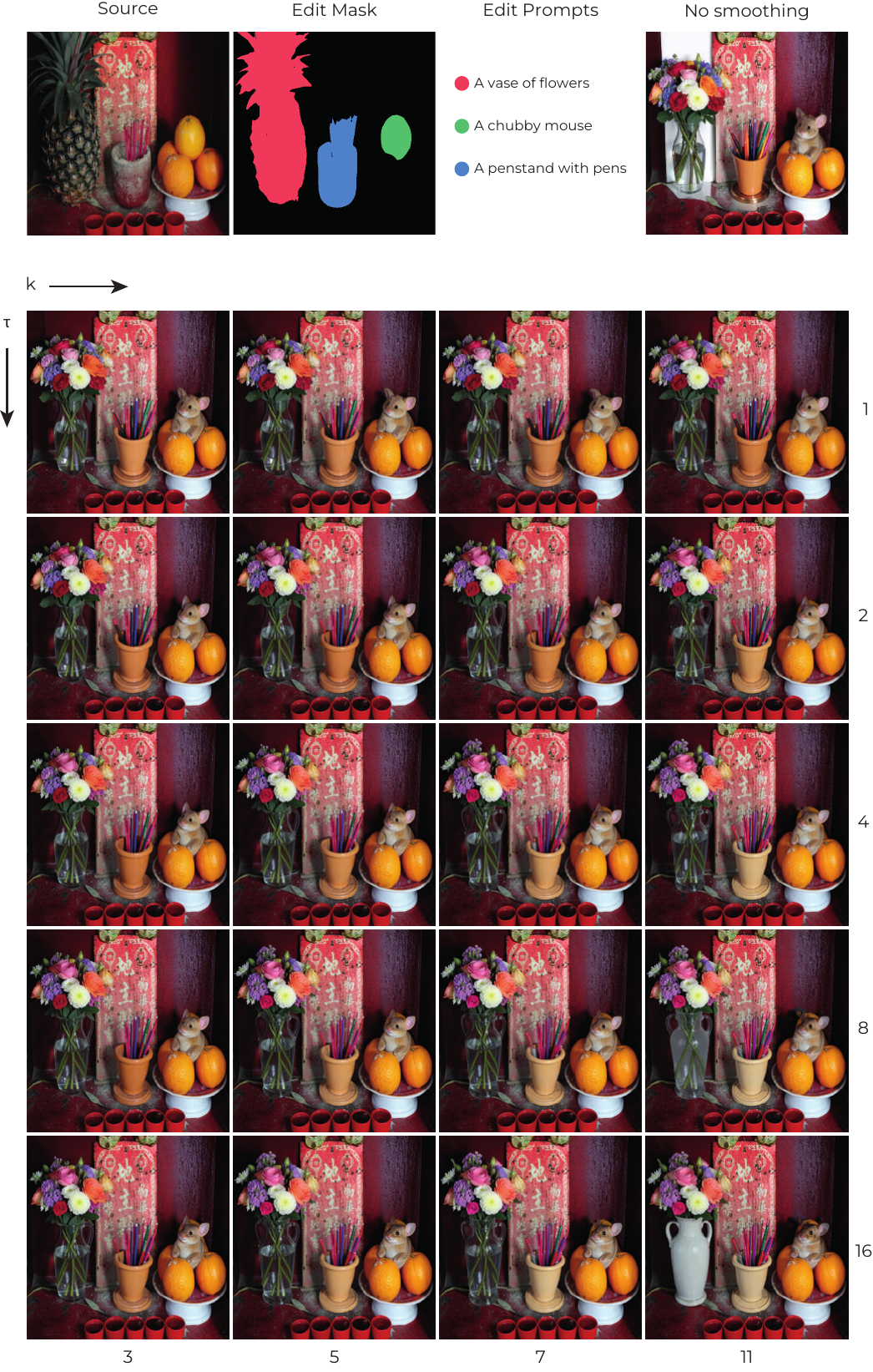}
    \caption{Qualitative effect of the kernel-size $k$ and temperature $\tau$ hyperparameters of MICE when applied to FLUX.2 [klein] 4B.}
    \label{fig:ablation}
\end{figure}

We report how the kernel size $k$ and the temperature $\tau$ affects the results in~\Cref{tab:full_ablation_kt}. The method is quite robust to hyperparameter choice. Considering all scores, the best overall configurations are $k=3,\tau=1$ for LoMOE-Bench and $k=3,\tau=8$ for MICE-Bench.
The effect of these hyperparameters is more clearly seen in the qualitative results for a sample from LoMOE-Bench shown in~\Cref{fig:ablation}, in which the pineapple on the left is to be replaced by a vase of flowers, which has a different shape. The version with no smoothing fails to harmonize with the background, drawing a white patch behind the vase. When smoothing the mask, different $k$ and $\tau$ lead to slightly different versions, with the vase converging to a similar style as the penstand on its right with increasing $k$ and $\tau$ values.

\begin{table*}[h]
\centering
\caption{Quantitative effect of the kernel-size $k$ and temperature $\tau$ hyperparameters of MICE when applied to FLUX.2 [klein] 4B.}\vspace{-.3cm}
\label{tab:full_ablation_kt}
\resizebox{\linewidth}{!}{%
\begin{tabular}{cc cccc cccc}
\toprule
& & \multicolumn{4}{c}{\textbf{LoMOE-Bench}} & \multicolumn{4}{c}{\textbf{MICE-Bench}} \\
\cmidrule(r{1.5em}){3-6} \cmidrule{7-10}
\textbf{$k$} & \textbf{$t$} & \textbf{Tgt C $\uparrow$} & \textbf{Loc C $\uparrow$} & \textbf{MAE$_\text{B}$ $\downarrow$} & \textbf{AR [\%] $\uparrow$} & \textbf{Tgt C $\uparrow$} & \textbf{Loc C $\uparrow$} & \textbf{MAE$_\text{B}$ $\downarrow$} & \textbf{AR [\%] $\uparrow$}\\
\midrule
1 & 1 & 25.78 & 29.59 & 7.60 & 96.35 & 24.95 & 26.63 & 8.15 & 95.36 \\
3 & 1 & 25.77 & 29.57 & 7.07 & 96.35 & 24.94 & 26.55 & 7.58 & 94.29 \\
5 & 1 & 25.67 & 29.53 & 6.98 & 96.35 & 24.90 & 26.47 & 7.41 & 93.95 \\
7 & 1 & 25.66 & 29.50 & 6.94 & 95.83 & 24.90 & 26.43 & 7.31 & 93.65 \\
11 & 1 & 25.67 & 29.50 & 6.87 & 95.83 & 24.85 & 26.37 & 7.16 & 93.07 \\
\midrule
3 & 2 & 25.63 & 29.46 & 6.91 & 96.35 & 24.94 & 26.55 & 7.47 & 94.29 \\
5 & 2 & 25.58 & 29.42 & 6.79 & 94.79 & 24.95 & 26.52 & 7.33 & 93.95 \\
7 & 2 & 25.55 & 29.41 & 6.72 & 94.79 & 25.05 & 26.55 & 7.27 & 94.00 \\
11 & 2 & 25.59 & 29.36 & 6.66 & 94.27 & 24.94 & 26.51 & 7.17 & 93.02 \\
\midrule
3 & 4 & 25.50 & 29.41 & 6.83 & 94.79 & 25.03 & 26.72 & 7.35 & 93.65 \\
5 & 4 & 25.62 & 29.33 & 6.64 & 94.27 & 25.04 & 26.68 & 7.30 & 93.31 \\
7 & 4 & 25.68 & 29.33 & 6.52 & 93.23 & 25.06 & 26.68 & 7.29 & 92.97 \\
11 & 4 & 25.71 & 29.35 & 6.43 & 92.71 & 25.06 & 26.68 & 7.30 & 92.77 \\
\midrule
3 & 8 & 25.57 & 29.35 & 6.37 & 92.71 & 25.15 & 26.85 & 7.84 & 92.48 \\
5 & 8 & 25.53 & 29.29 & 6.30 & 91.15 & 25.14 & 26.83 & 7.85 & 92.24 \\
7 & 8 & 25.55 & 29.28 & 6.26 & 91.15 & 25.15 & 26.83 & 7.86 & 91.80 \\
11 & 8 & 25.58 & 29.27 & 6.24 & 91.15 & 25.11 & 26.80 & 7.92 & 91.60 \\
\midrule
3 & 16 & 25.52 & 29.32 & 6.23 & 90.62 & 25.15 & 26.71 & 9.05 & 91.85 \\
5 & 16 & 25.55 & 29.24 & 6.21 & 90.62 & 25.15 & 26.69 & 9.07 & 91.85 \\
7 & 16 & 25.53 & 29.20 & 6.19 & 90.62 & 25.15 & 26.69 & 9.10 & 92.09 \\
11 & 16 & 25.55 & 29.18 & 6.15 & 90.62 & 25.14 & 26.66 & 9.15 & 92.19 \\
\bottomrule
\end{tabular}%
}\vspace{.1cm}
\end{table*}
\begin{figure}[h!]
    \centering
    \includegraphics[width=.99\linewidth]{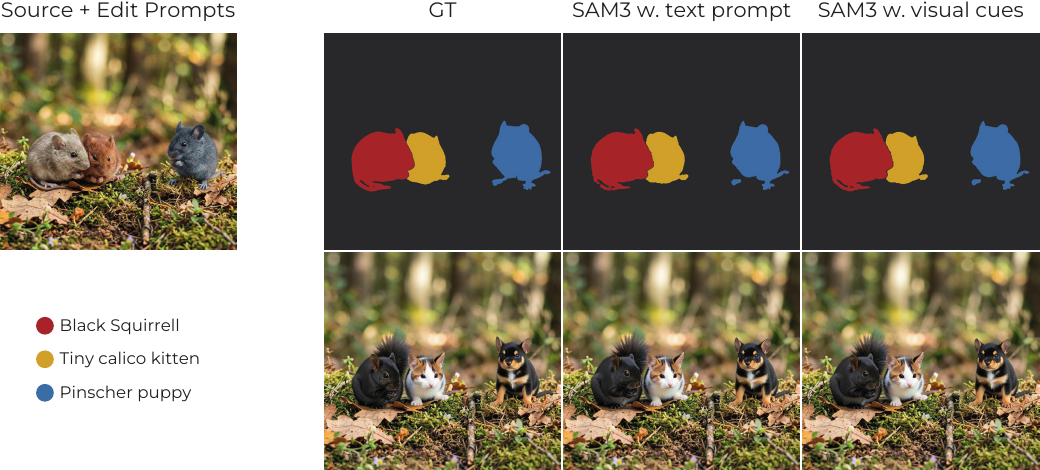}
    \vspace{-.2cm}
    \caption{Qualitative comparison of feeding MICE with instance segmentation masks obtained by prompting SAM3 with text prompts or visual prompts (dots on the centroid of the instance), and with the ground truth masks.}
    \label{fig:sam3_ablation}
\end{figure}

\section{Combination with Automatically-Obtained Instance Masks}
Our approach takes as editing instructions some textual prompts and segmentation masks, localizing each instance to be edited. Such masks can be directly provided by the user or determined with an external promptable instance segmentation model. As an example, we consider feeding MICE with masks obtained by prompting the state-of-the-art SAM3~\cite{carion2025sam} instance segmentation approach. Specifically, we prompt SAM3 with the target instance information contained in the editing prompt (\eg~\textit{the gray mouse on the left}; \textit{the chestnut mouse in the center}; \textit{the dark gray mouse on the right}) or with visual cues in the form of points in the center of the instance. We argue that both these strategies are feasible in practice, making our approach easy to use. More importantly, even when fed with non-perfect masks, our approach leads to consistent results (see~\cref{fig:sam3_ablation}), thanks to the smoothing it entails.

\section{Qualitative Analysis of Different Editing Areas Definition Strategies}
\begin{figure}
    \centering
    \includegraphics[width=\linewidth]{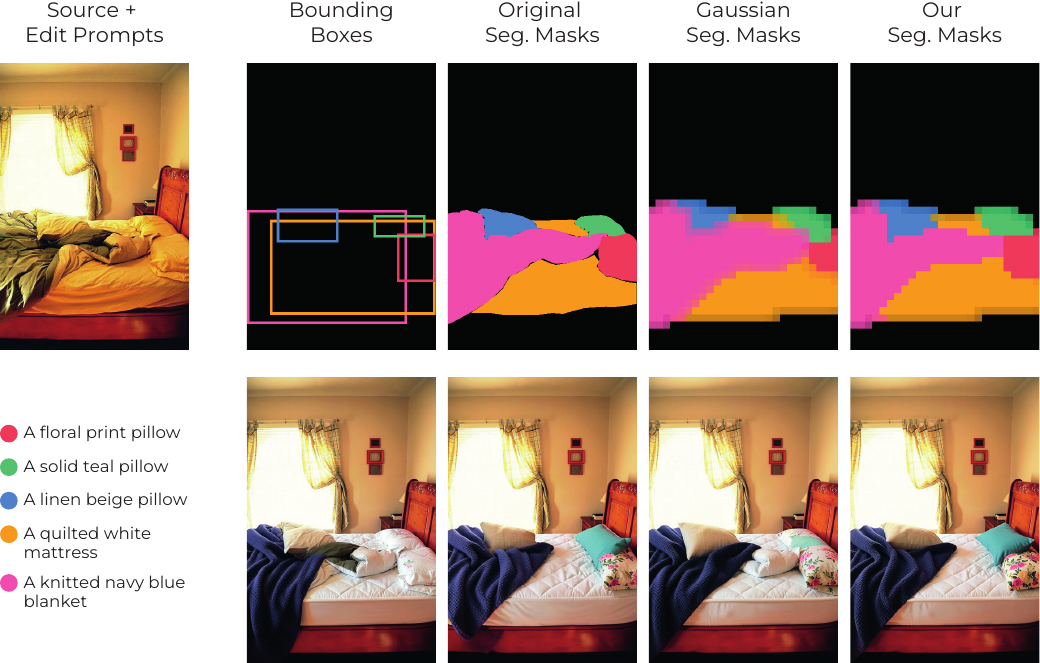}
    \caption{Qualitative results obtained when using MICE in combination with different strategies to localize the instances to edit.}
    \label{fig:instance_loc}
\end{figure}

Our approach adapts the attention bias for the tokens relative to each instance to edit. The visual tokens are determined starting from modified segmentation masks of the instances. Our modification entails applying a Gaussian smoothing kernel to the original segmentation masks, making sure that the smoothing does not cause the masks to overlap with each other. An alternative strategy adopted in literature~\cite{zaccagnino2026shifting, zhou2025dreamrenderer} consists in using bounding boxes instead of segmentation masks. In~\cref{fig:instance_loc}, we report the qualitative effect of alternative strategies to localize the instances. As we can observe, when bounding boxes overlap, the model struggles at executing all the edits. When using the original segmentation masks, the shape of the instances is better preserved, but some artifacts can remain due to poor harmonization between instances (see \eg~the teal and floral pillows). By simply smoothing the masks with the Gaussian kernel, some instances are more likely to change shape and be too blended together. On the other hand, our instance-aware Gaussian smoothing strategy helps following all the editing prompts while preserving the original shape of the instances when possible, and allows better harmonization between the instances and the entire image.

\section{Prompting Strategies for Multi-Instance Editing}
To assess the benefits of our proposed approach for multi-instance image editing, we apply it to the state-of-the-art FLUX.2 [klein] FM-based generative model, in its 4B configuration. We determine the best strategy to obtain a fair baseline by prompting the model as proposed by its authors. Specifically, we compare the vanilla usage of FLUX.2 [klein], where a single global editing instruction is provided (Text Loc. in the table). A single overall edit prompt including all instances and localization information is provided for both LoMOE-Bench and MICE-Bench, so we use that. To give the same information to the baseline model that we give to our model (edit prompts in the format \texttt{Replace the \{SRC\} with \{TGT\}} and visual reference), we also add visual cues to the image. Since, as shown in the second row of~\Cref{tab:prompting_flux} (where \textbf{Vis. Loc.} is enabled), this degrades image quality significantly, we also exploited FLUX.2's ability to use multiple input images to provide two images as context, so we provide one with visual cues and one without. This degrades prompt following ability even further, as reported by the third row of ~\Cref{tab:prompting_flux} (where $\mu$Ref is enabled).

As a reference, we also report the performance of FLUX.2 [klein] when used in a multi-turn setting ($\mu$Turn in the table), where the editing process is decomposed into multiple sequential turns, each applying an editing instruction conditioned on the output of the previous step, both when using just the source image as context, with visual localization, and in the multi-reference setting. The $\mu$Turn variants bring advantages on the simpler LoMOE-Bench, but their performance deteriorates as the number of requested edits increases (as happens in MICE-Bench). In light of these results, the best strategy is the plain end-to-end version without any added visual localization, which requires prompts with localization information. Therefore, we use this specific prompting strategy whenever we refer to the FLUX.2 baseline models, to Gemini 3 Pro Image or to Qwen-Image-Edit.

\begin{table*}[t]
\centering
\caption{Quantitative comparison of baseline models with different prompting strategies on LoMOE and MICE-Bench. We refer to the multi-turn setting with $\mu$Turn, to the two input context image setting with $\mu$Ref, and to the visual cues setting with Vis. Loc.}\vspace{-.2cm}
\label{tab:prompting_flux}
\resizebox{\linewidth}{!}{%
\begin{tabular}{c ccc @{\hskip 1.5em} cccc @{\hskip 1.5em} cccc}
\toprule
& & & & \multicolumn{4}{c}{\textbf{LoMOE}} & \multicolumn{4}{c}{\textbf{MICE-Bench}} \\
\cmidrule(r{1.5em}){5-8} \cmidrule{9-12}
\textbf{$\mu$Turn} & \makecell[c]{\textbf{$\mu$Ref}} & \makecell[c]{\textbf{Text}\\ \textbf{Loc.}} & \makecell[c]{\textbf{Vis.}\\ \textbf{Loc.}} & \textbf{Tgt C $\uparrow$} & \textbf{Loc C $\uparrow$} & \textbf{MAE$_\text{B}$ $\downarrow$} & \textbf{AR [\%] $\uparrow$} & \textbf{Tgt C $\uparrow$} & \textbf{Loc C $\uparrow$} & \textbf{MAE$_\text{B}$ $\downarrow$} & \textbf{AR [\%] $\uparrow$} \\
\midrule
& & \checkmark & & 25.94 & 28.68 & 8.76 & 97.40 & 25.37 & 25.97 & 26.08 & 99.61 \\
&  &  & \checkmark & 23.38 & 27.97 & 7.78 & 92.71 & 24.14 & 24.16 & 21.48 & 99.71 \\
 & \checkmark &  & \checkmark & 23.26 & 27.66 & 5.94 & 78.12 & 23.99 & 24.02 & 21.54 & 97.46 \\
\midrule
\checkmark &  &  &  & 25.49 & 28.73 & 13.81 & 98.96 & 24.75 & 24.29 & 29.65 & 99.95 \\
\checkmark &  & \checkmark &  & 25.61 & 28.44 & 12.17 & 95.83 & 25.21 & 25.17 & 29.97 & 100.00 \\
\checkmark &  &  & \checkmark & 24.38 & 28.11 & 12.31 & 97.40 & 24.26 & 24.21 & 29.48 & 99.95 \\
\checkmark &  & \checkmark & \checkmark & 24.51 & 28.24 & 11.40 & 94.27 & 25.24 & 25.19 & 29.28 & 100.00 \\
\checkmark & \checkmark &  & \checkmark & 24.72 & 28.23 & 11.45 & 95.83 & 24.82 & 24.04 & 31.12 & 100.00 \\
\checkmark & \checkmark & \checkmark & \checkmark & 25.17 & 28.06 & 10.53 & 92.19 & 25.43 & 24.70 & 31.35 & 100.00 \\
\bottomrule
\end{tabular}%
}
\end{table*}

\begin{figure}[h]
    \centering
    \includegraphics[width=0.85\linewidth]{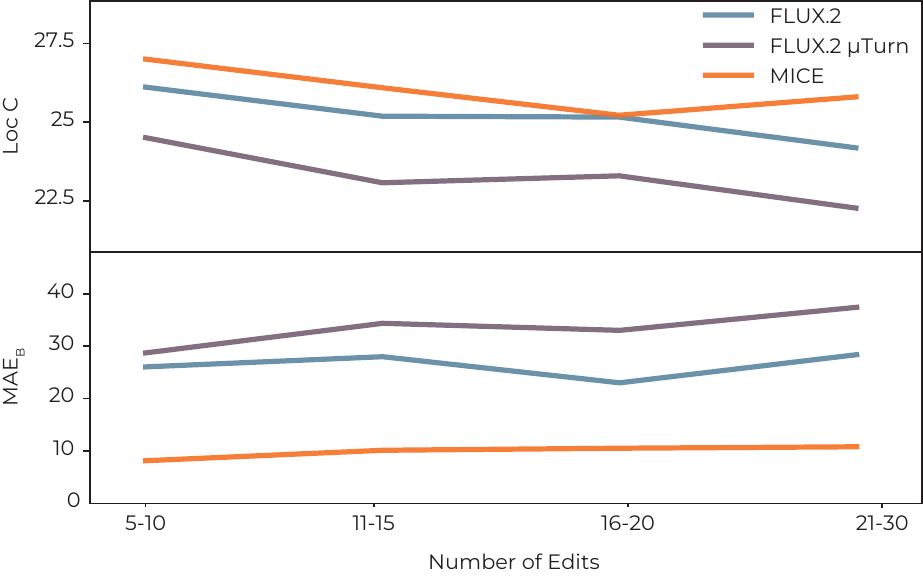}
    \caption{Visual quality of generated images with increasing number of edits.}
    \label{fig:scores_scaling_discussion}
\end{figure}

Moreover, in~\Cref{fig:scores_scaling_discussion} we see how the visual quality of the generated images scales with increasing number of edits using the baseline FLUX.2 [klein] model in single-turn, multi-turn, and with the addition of MICE. As we can see, both single-turn and multi-turn baselines show significant degradation of visual quality, due to ignored edit instructions and error accumulation, while the proposed MICE scales more gracefully.

\section{Qualitative Results of Adaptation Strategies for Multi-Instance Concurrent Editing}

\begin{figure}
    \centering
    \includegraphics[width=\linewidth]{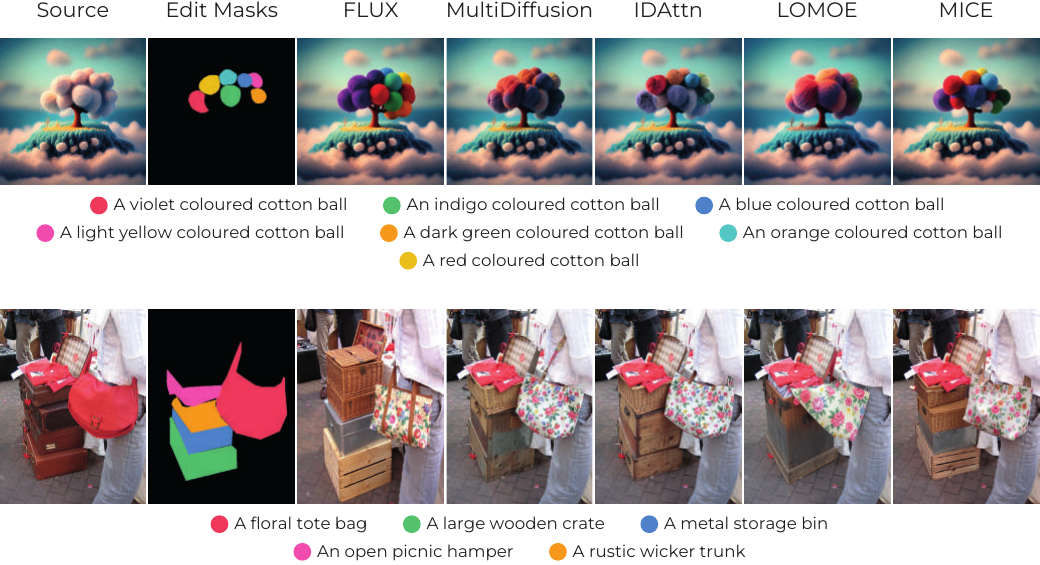}
    \caption{Porting on the same FLUX.2 [klein]-4B baseline of competitors on LoMOE-Bench (first row) and MICE-Bench (second row).}\vspace{-0.5cm}
    \label{fig:mice_strategies}
\end{figure}

In~\Cref{fig:mice_strategies} we report a qualitative comparison on a sample from each dataset of various strategies for Multi-Instance Concurrent Editing, all running on the FLUX.2 [klein] 4B baseline. FLUX, which is the baseline model, fails on both samples to properly follow the prompts, MultiDiffusion gets closer, but it merges some object together on both samples (some of the cotton balls in the LoMOE-Bench sample, almost all of the stacked containers in the MICE-Bench sample). LoMOE performs even worse in this regard, creating very big merged objects from the original ones in both samples. IDAttn does not exhibit this destructive behavior to such an extent, but still fails to properly follow all prompts. MICE is able to follow the prompts, but in the cotton ball sample one of the cotton balls which should not be modified is instead recolored.

\section{Qualitative Results on the Backbone Adaptability} 
\begin{figure}[h]
    \centering
    \includegraphics[width=\linewidth]{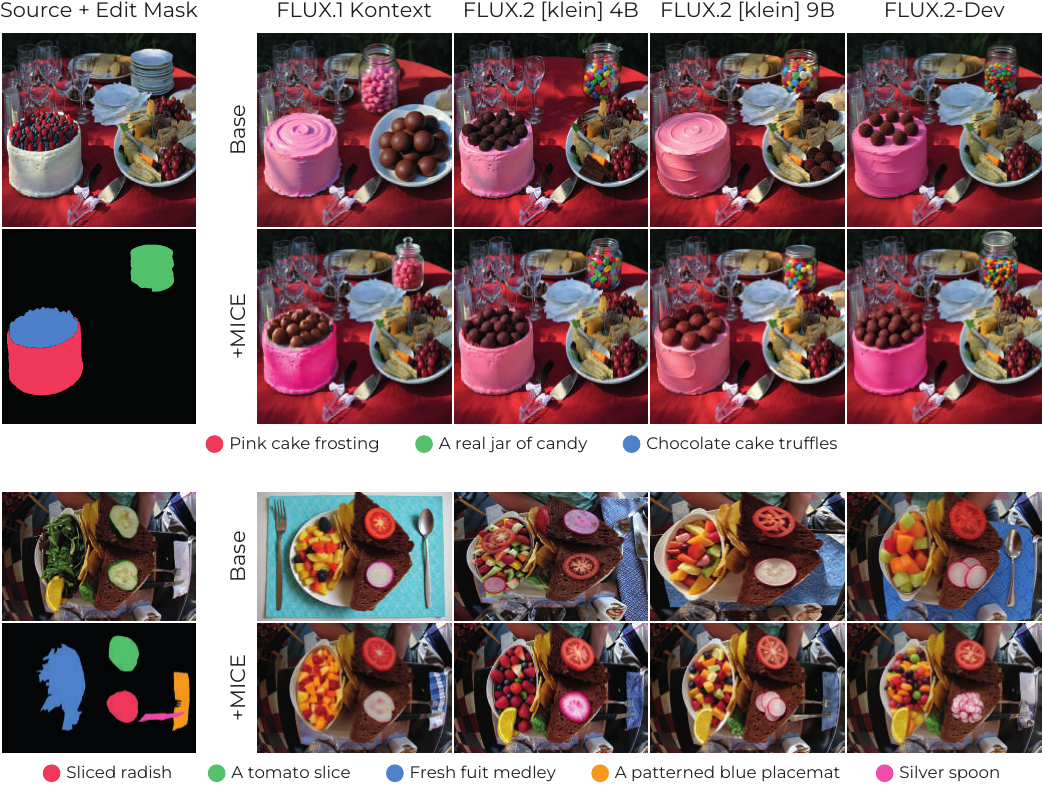}
    \caption{Qualitative results of applying MICE to different backbones on LoMOE-Bench (first row) and MICE-Bench (second row).}
    \label{fig:porting}
\end{figure}

Our inference-time approach for multi-instance concurrent editing can be applied seamlessly to the MMDiT of different generative editing models. The qualitative effect of applying MICE to architectures of different sizes and input processing pipelines is showcased in~\cref{fig:porting}. It can be observed that, despite the visual quality of the final output being determined by the capabilities of the backbone, the adherence to the prompts in terms of semantics and locality is consistently ensured by MICE.

\begin{table}[h!]
    \centering
    \caption{User Study Results (ELO scores).}\vspace{-0.2cm}
    \label{tab:user_study}
    \resizebox{0.55\linewidth}{!}{%
        \begin{tabular}{l c c}
        \toprule
         & \textbf{LoMOE-Bench} & \textbf{MICE-Bench} \\
        \midrule
        MICE                & 1335.48 & 1406.00 \\
        IDAttn              & - & 1018.80 \\
        LoMOE               & 1080.71 & -  \\
        FLUX.2 [klein]      & 1183.80 & 1175.20 \\
        \bottomrule
        \end{tabular}%
    }
\end{table}

\section{User Study}\label{sec:user_study}

To complement the analysis performed in Section 5, we perform a user study, asking 30 people to answer 60 binary-choice questions choosing which among two proposed generated images, chosen among the ouputs of FLUX.2 [klein] 9B, MICE-9B and LoMOE for 10 samples of LoMOE-Bench, and FLUX.2 [klein] 9B, MICE-9B and IDAttn for 10 samples of MICE-Bench is more faithful to the given editing prompt. We compute ELO scores from these responses and report them in~\Cref{tab:user_study}.
Despite its high CLIP Score, LoMOE underpeforms when evaluated by humans or VLMs, as noted also by~\cite{zaccagnino2026shifting}. MICE outperforms all competitors on both datasets.

\section{Remarks On the Quantitative Evaluation}
\begin{figure}[h!]
    \centering
    \includegraphics[width=0.88\linewidth]{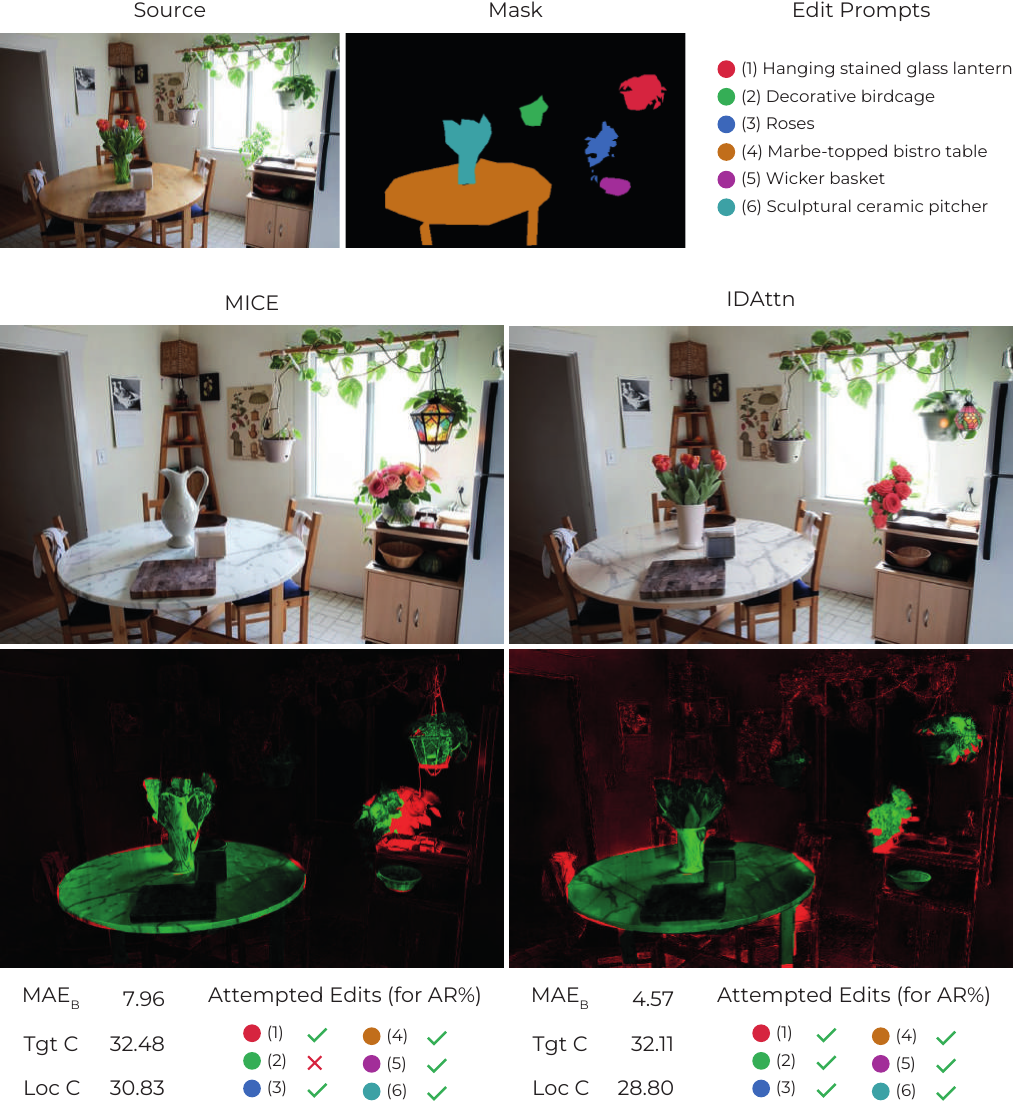}\vspace{-0.3cm}
    \caption{Scores and error analysis of MICE and IDAttn with pixel-wise difference \wrt~the source (green contributes to AR\%, red to MAE$_\text{B}$).}
    \label{fig:scores_discussion}
\end{figure}

Evaluating generative tasks, including the multi-instance generative editing one at hand, is notably tricky. Therefore, it is preferable to compare approaches based on human preference (see~\Cref{sec:user_study}) or a proxy LLM-as-judge analysis, as the one reported in the main paper. Nonetheless, quantitative scores are useful to underline specific behaviors of the model when interpreted correctly. To give an example, in~\Cref{fig:scores_discussion} we compare the generated images using our proposed MICE and IDAttn, showing what the scores used focus on. For example, the MAE$_\text{B}$ is higher in the MICE sample than in the IDAttn sample due to the harmonization and more edits performed. Similarly, heuristics such as AR\%, while useful in comparing the number of attempted edits, are also affected by false positives (such as edits 6 and 2 from IDAttn). As advised by its authors~\cite{zaccagnino2026shifting}, this score should be used as a sanity check rather than for model ranking.\vspace{-0.3cm}

\section{Further Analysis of the Inference Time}
In this section, we compare the inference time behavior of our MICE method compared to a the multi-turn version of FLUX.2 [klein], to MultiDiffusion, to LoMOE (only theoretically), and to IDAttn, which are respectively a representative of a multi-turn approach, two multi-branch approaches (w/w.o. latent optimization) and an end-to-end method.

\begin{figure}[t] %
    \centering
    \resizebox{\textwidth}{!}{
    \begin{minipage}{\textwidth}
        \centering
        \includegraphics[width=0.495\linewidth]{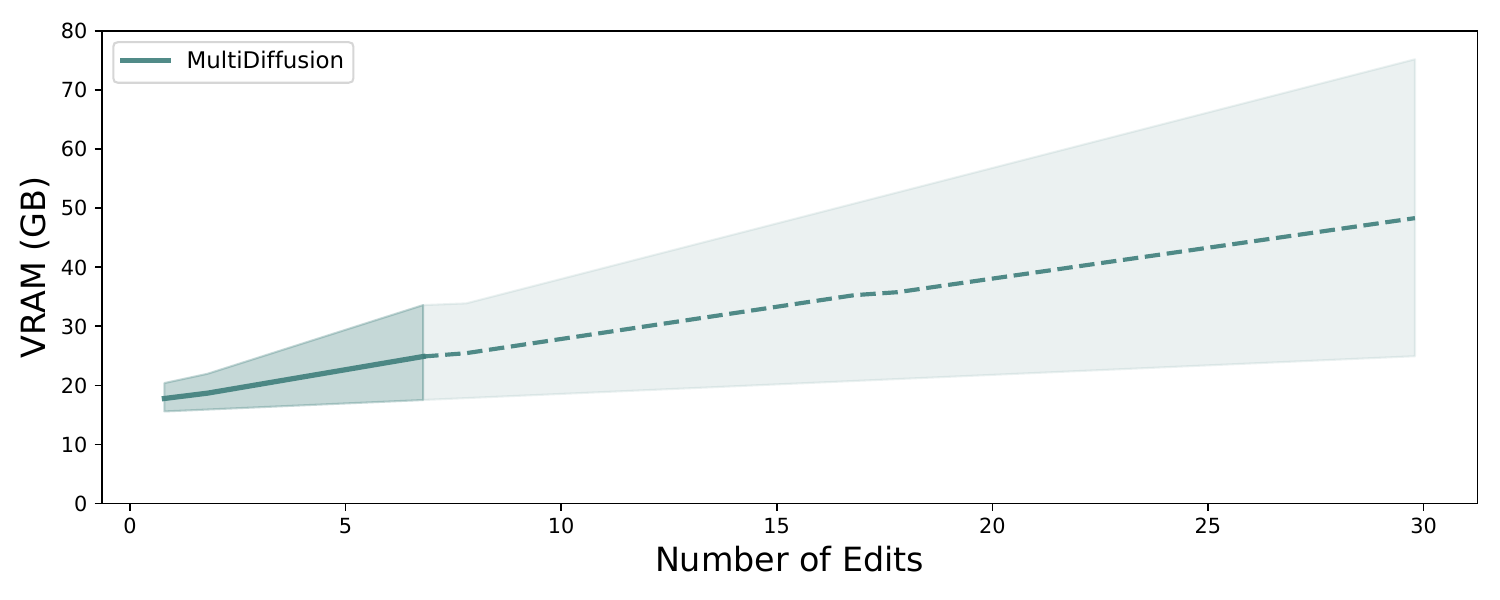}\hfill
        \includegraphics[width=0.495\linewidth]{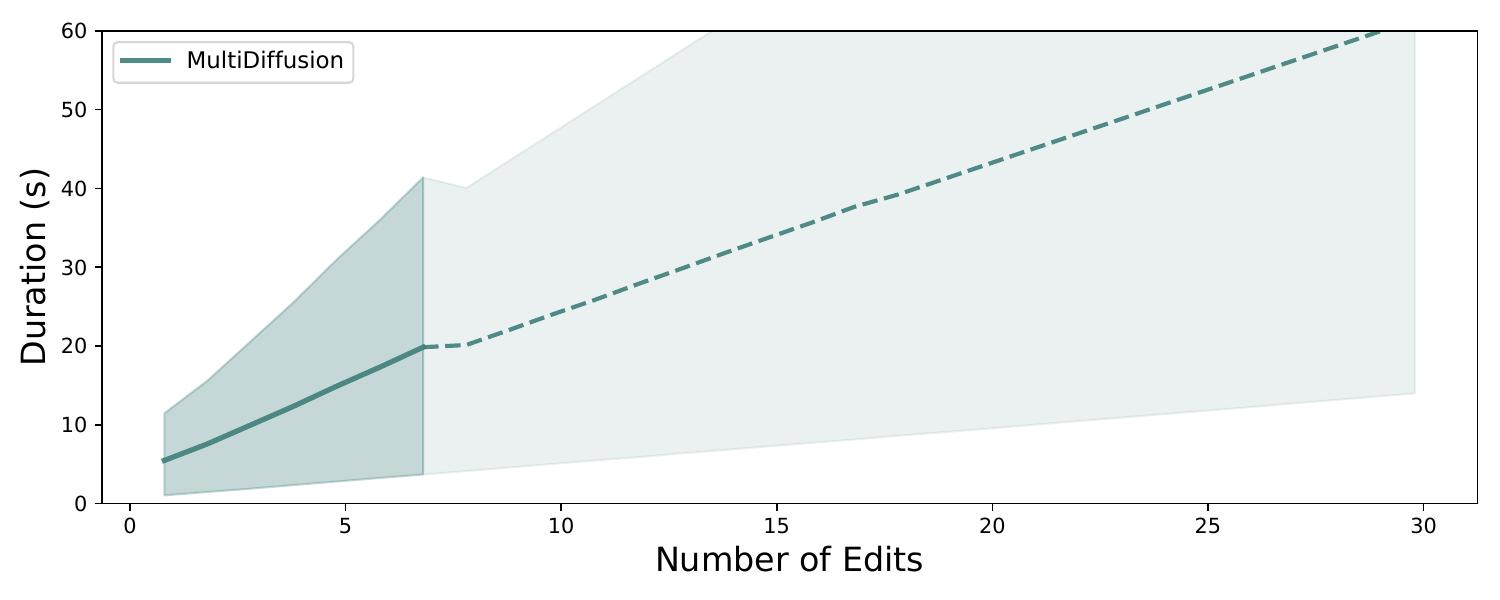}
        \includegraphics[width=0.495\linewidth]{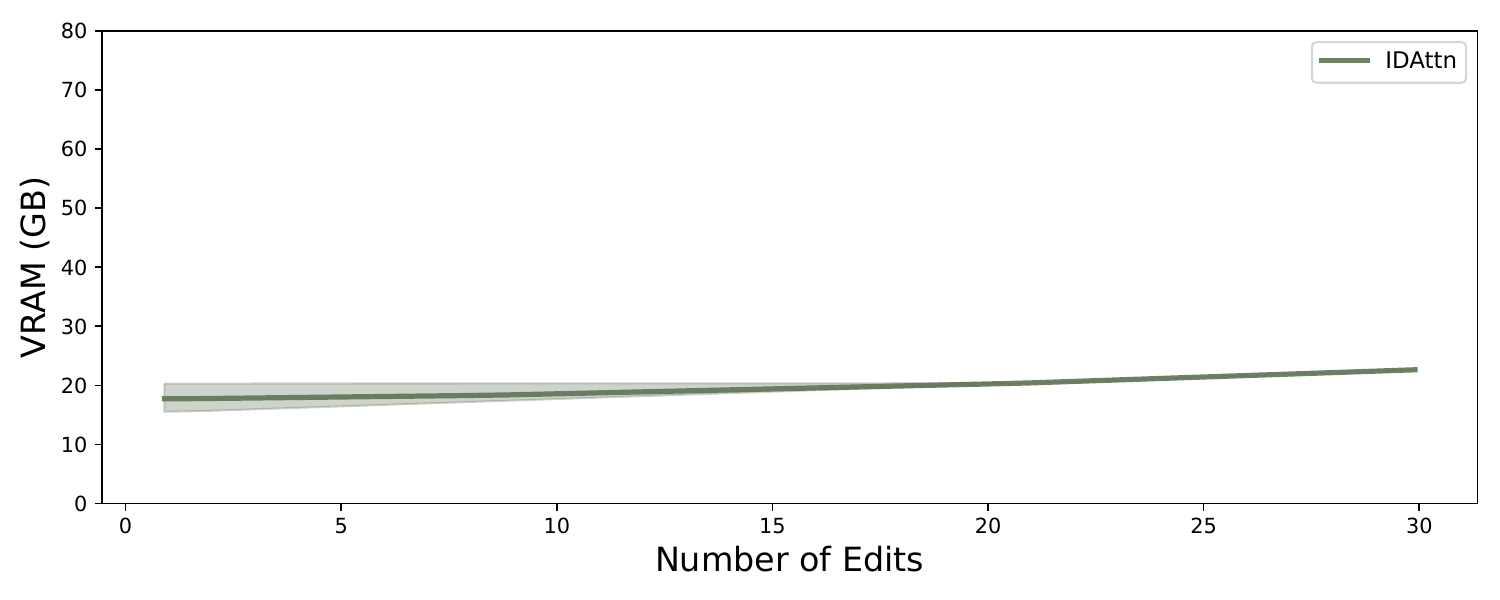}\hfill
        \includegraphics[width=0.495\linewidth]{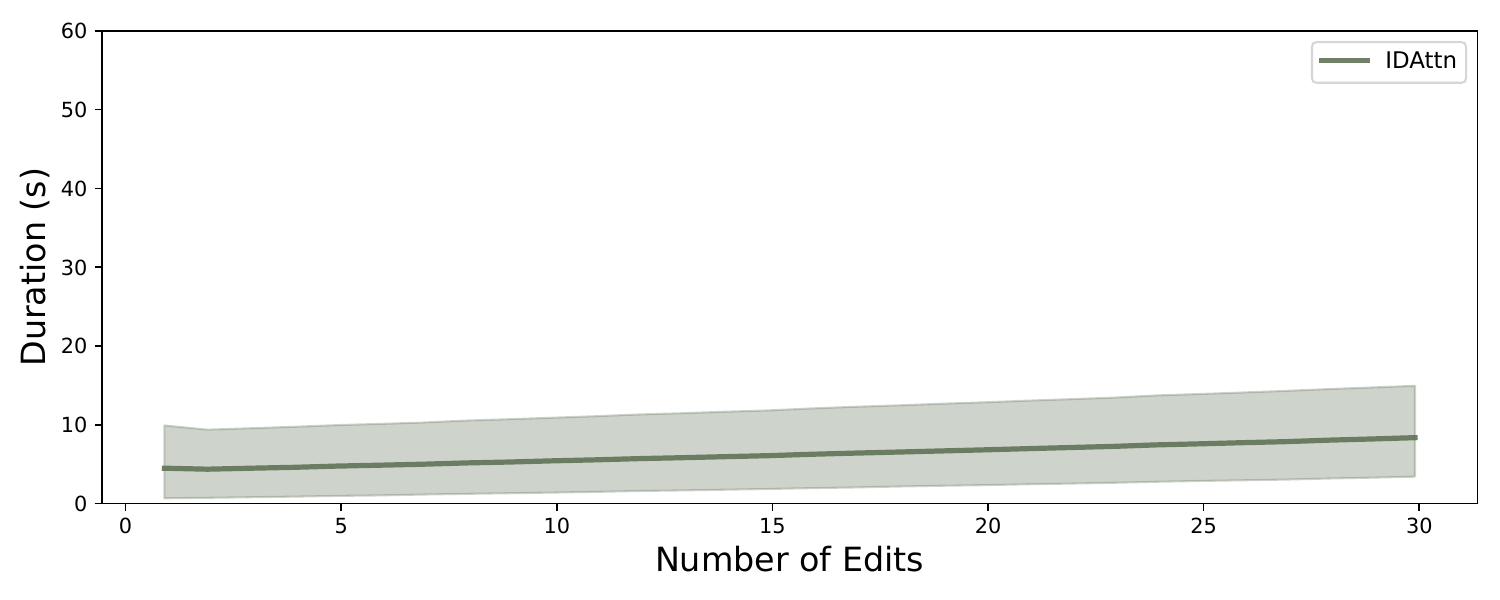}
        \includegraphics[width=0.495\linewidth]{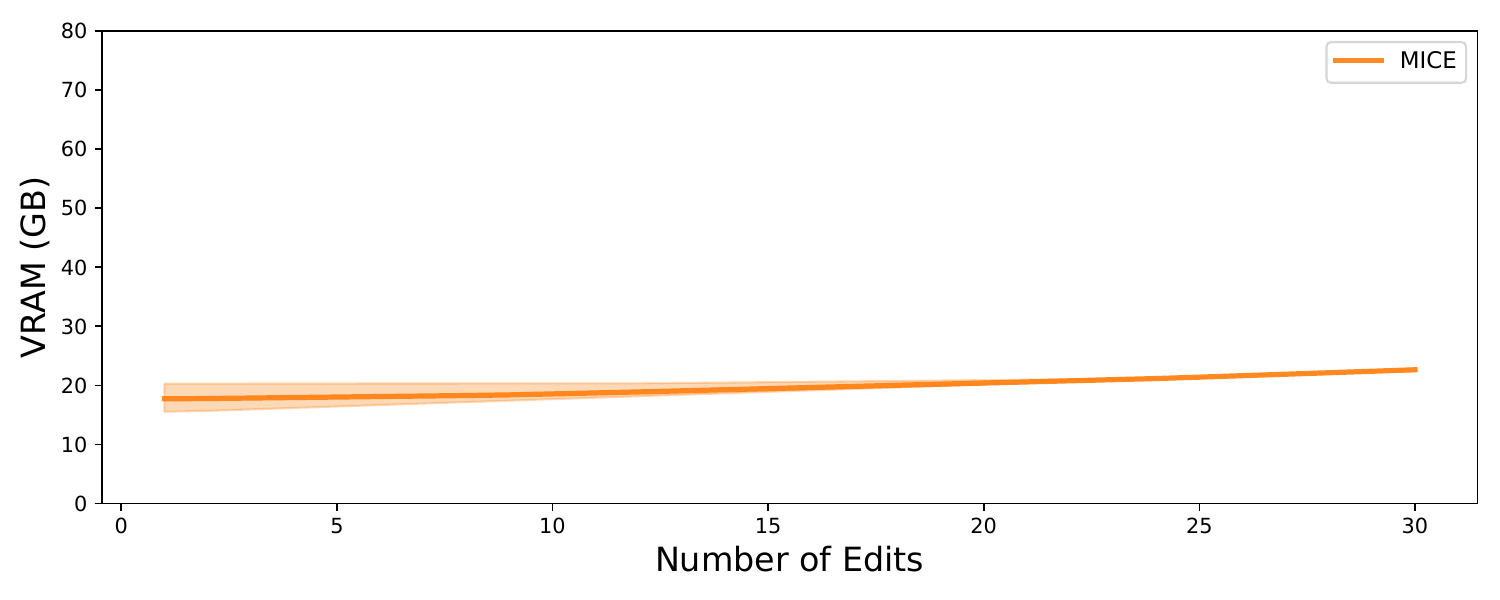}\hfill
        \includegraphics[width=0.495\linewidth]{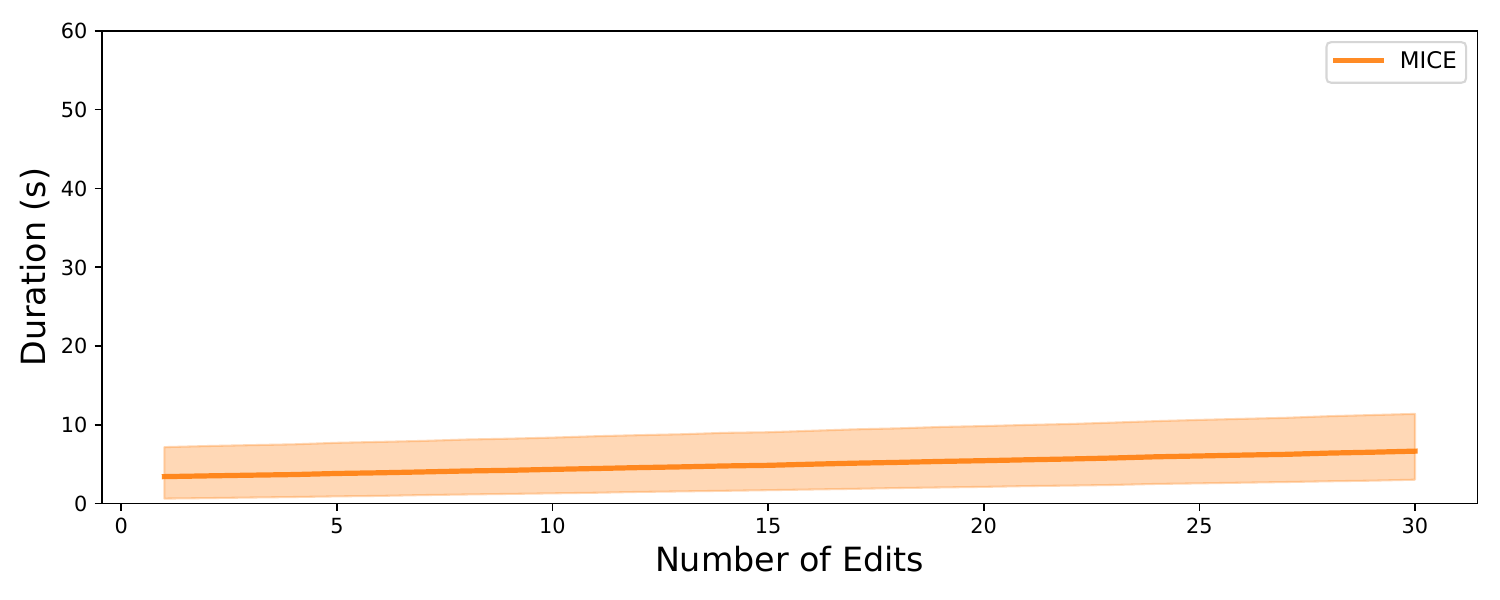}
        \includegraphics[width=0.495\linewidth]{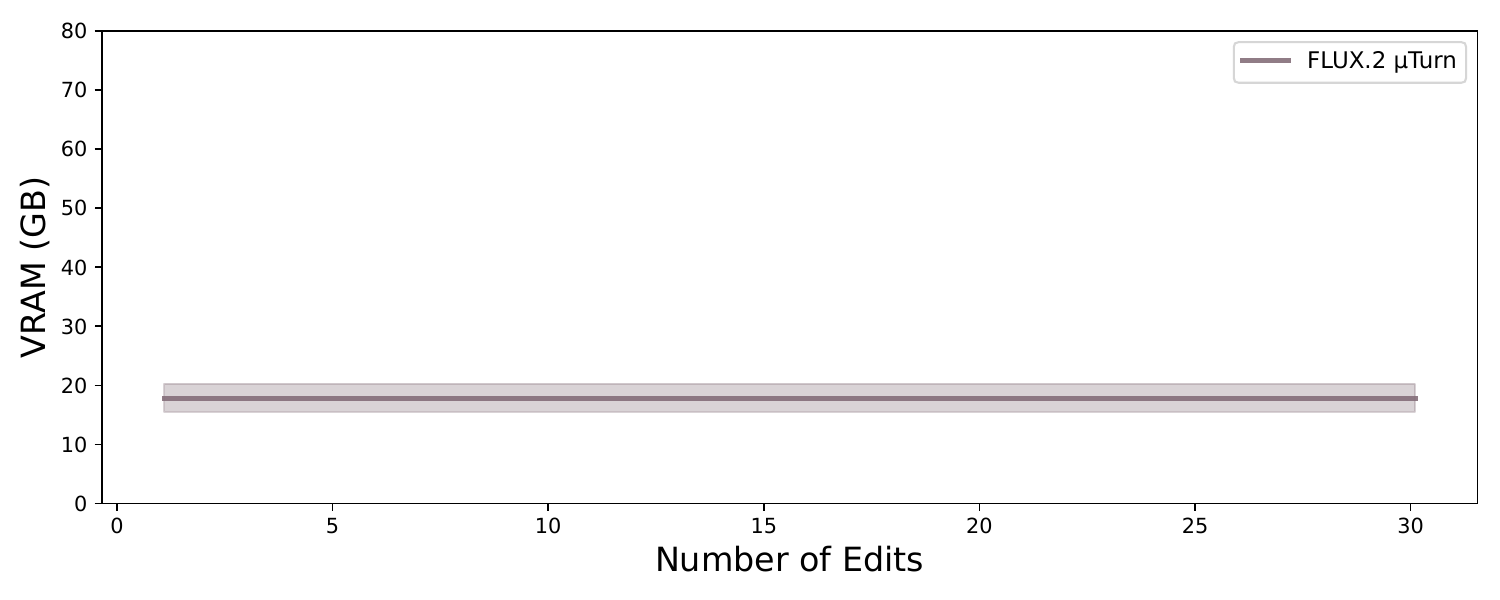}\hfill
        \includegraphics[width=0.495\linewidth]{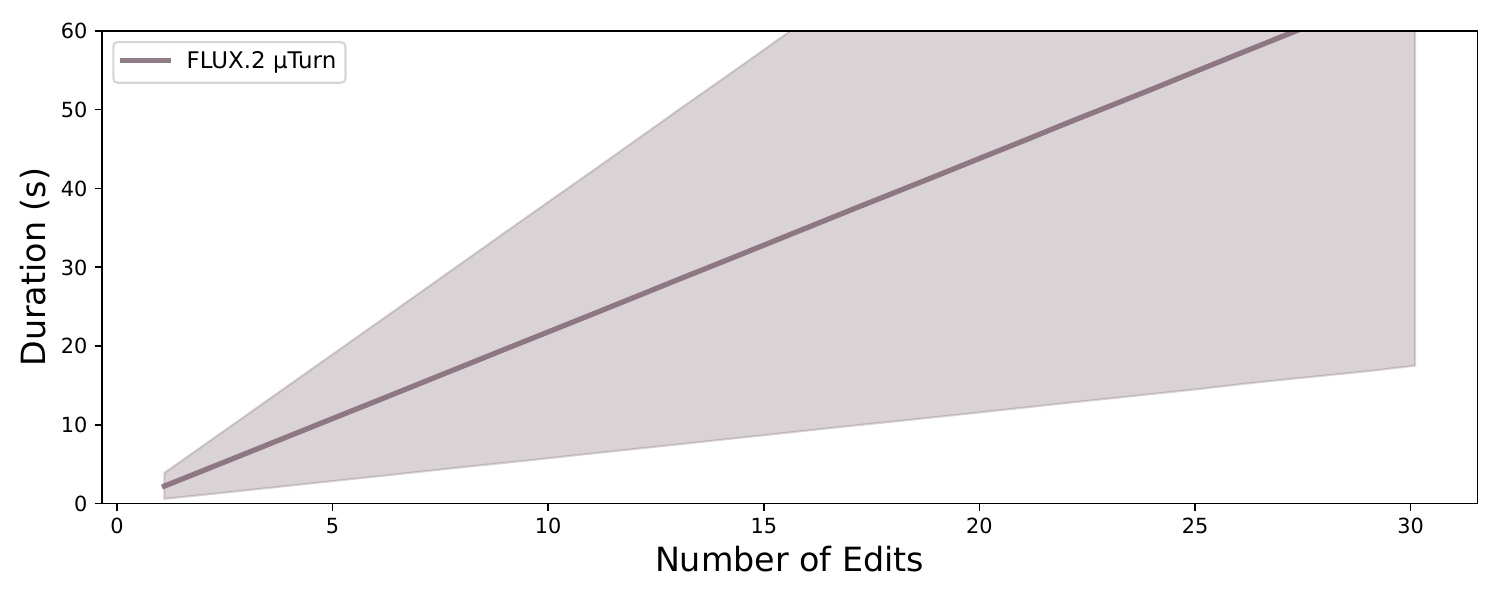}
        \includegraphics[width=0.495\linewidth]{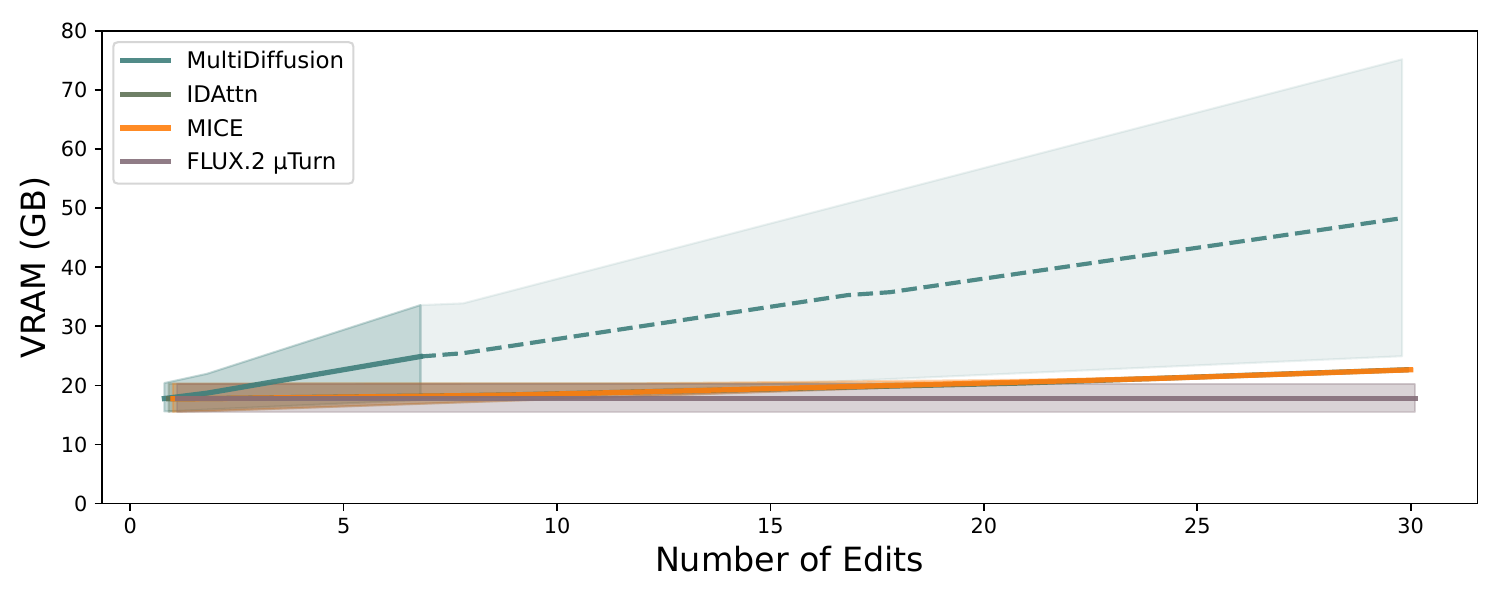}\hfill
        \includegraphics[width=0.495\linewidth]{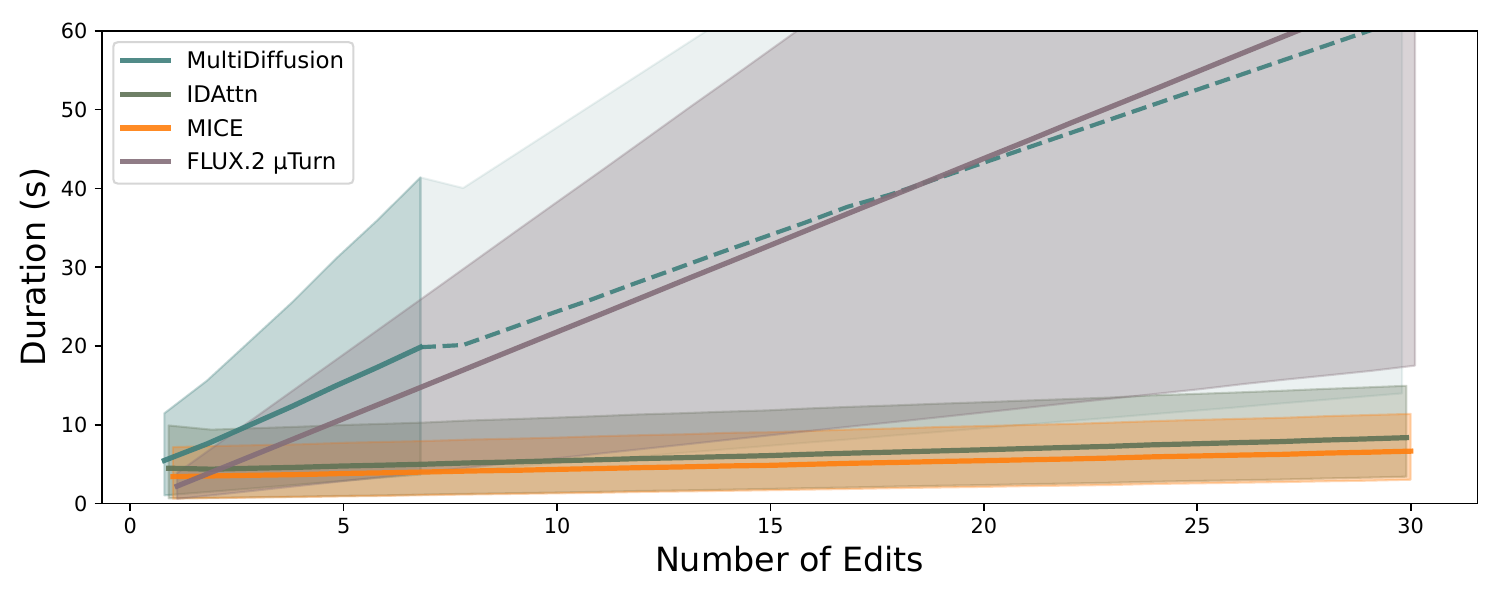}
        \captionof{figure}{VRAM utilization and inference duration \wrt~the number of edits.}
        \label{fig:vram_time_scaling_all}
    \end{minipage}%
    }
\end{figure}

\tit{Empirical evaluation}
First, we perform an empirical analysis by benchmarking all methods on the same samples. We extrapolate values using least-squares when methods require more than our GPU memory budget and indicate it via dashed lines. We show results in~\Cref{fig:vram_time_scaling_all} using a 40GB NVIDIA A100 GPU. We test the methods at resolution $1024\times 1024$ and represent the performance using lines, and also show variance by benchmarking at resolutions $512\times 512$ and $1536\times 1536$, represented as bands in the charts.  As we can see, both multi-turn approaches and multi-branch approaches are not practical when number of edits increases, as they iterate on sequential model calls (increasing time) or parallel model calls (increasing VRAM usage). We didn't include LoMOE in this empirical evaluation because, to avoid running out of VRAM, we need to enable gradient checkpointing and other such modifications, which would make this comparison unfair. Nonetheless, the following theoretical analysis shows its overhead.

\tit{Theoretical analysis}
Here we elaborate on the inference-time efficiency differences between the considered approaches with a theoretical analysis of their computational and memory complexity.
Let $N$ denote the number of instance edits and $B$ the number of transformer blocks. 
Each MMDiT forward pass performs attention over a combined sequence of text tokens ($\mathbf{Z}^\text{prompt}$) and image tokens ($\mathbf{Z}^\text{img} = \mathbf{Z}^\text{latent} \parallel \mathbf{Z}^\text{context}$).
We define $\mathcal{F}(\mathbf{|Z|})$ as the cost of one full transformer forward pass, whose cost scales as $\mathcal{O}(\mathbf{|Z|}^2)$ (quadratic with respect to the total sequence length), where $\mathbf{Z} = \mathbf{Z}^\text{prompt} \parallel \mathbf{Z}^\text{latent} \parallel \mathbf{Z}^\text{context}$. The possible strategies for performing multi-instance concurrent editing, either designed or adapted for the task, have the following cost (also summarized in~\Cref{tab:complexity_asymptotic}):

\begin{itemize}
    \item\textbf{$\mu$Turn (Sequential).} The model is called $N$ times independently, once for each instance edit, applied sequentially to the image. Therefore, since the time per step is $N \cdot \mathcal{F}(\mathbf{|Z|})$, the total time for an inference run is $\mathcal{O}(N~\cdot~\mathbf{|Z|}^2)$. On the other hand, since the VRAM usage is determined by a single forward pass, as each call is independent. Thus, the total VRAM usage during inference is $\mathcal{O}(\mathbf{|Z|}^2)$.

    \item\textbf{MultiDiffusion (Multi-branch).} At each denoising step, the model is run once for each of the $N$ prompts and once for the background, denoising the full latent image and producing $N$+1 noise predictions. These predictions are then fused via weighted averaging according to the masks. For simplicity, we omit the background branch in the scaling computation. Then, the time per step is $N \cdot \mathcal{F}(\mathbf{|Z|})$. However, one can note that the $N$ branches could be executed in parallel since there are no dependencies between them, so the execution time can theoretically be reduced to $\mathcal{F}(\mathbf{|Z|})$. As for the VRAM, which is determined by the branches, each denoising its own copy of latents, its complexity is $\mathcal{O}(\mathbf{|Z|}^2 + N \cdot \mathbf{|Z|})$.
    
    \item\textbf{LoMOE (Multi-Branch with latent optimization).}
    Recall that LoMOE extends MultiDiffusion with a three-pass optimization at each timestep: a reconstruct phase to denoise the background, then $N$ prediction branches to compute cross-attention and segmentation losses, and a final denoising with optimized latents. We consider the cost of the backpropagation pass to be $2N$~\cite{pau2024mathematical}. Therefore, the time per step is $\approx 5N \cdot \mathcal{F}(\mathbf{|Z|})$. As for the VRAM usage, we consider the cost of keeping the gradients for all latents in memory at the output of each block, which is $\mathcal{O}(N \cdot B \cdot \mathbf{Z}^2)$.

    \item\textbf{IDAttn (End-to-End, multiple prompts).} This method requires a single forward pass through the model. The text tokens $\mathbf{Z}^\text{prompt}$ are divided into a fixed-lenght utility global prompt $\mathbf{Z}^\text{prompt}_g$ and $N$ instance prompts $\mathbf{Z}^\text{prompt}_n$, they are all concatenated into a single text sequence to obtain $\mathbf{Z}^\text{prompt}$. Therefore, the computation time and memory usage are dominated by the number of additional text tokens for the instances's prompts, which grow linearly with $N$. Therefore. In other words, each step takes $\mathcal{F}(|\mathbf{Z}^\text{img}| + N \cdot |\mathbf{Z}^\text{prompt}_n|)$, and the VRAM usage of the model follows $\mathcal{O}((|\mathbf{Z}^\text{img}| + N \cdot |\mathbf{Z}^\text{prompt}_n|)^2)$.

    \item\textbf{MICE (End-to-End, multiple prompts).} Our method uses a single-forward-pass and treats the text prompts $\mathbf{Z}^\text{prompt}$ similarly to IDAttn. Therefore, its required time per step is $\mathcal{F}(|\mathbf{Z}^\text{img}| + N \cdot |\mathbf{Z}^\text{prompt}_n|)$, and the VRAM occupied by the model scales as $\mathcal{O}((|\mathbf{Z}^\text{img}| + N \cdot |\mathbf{Z}^\text{prompt}_n|)^2)$.

\end{itemize}

\begin{table*}[t]
\centering
\caption{Asymptotic complexity of the considered strategies designed or adapted for multi-instance editing.}
\label{tab:complexity_asymptotic}
\resizebox{\linewidth}{!}{%
\begin{tabular}{l @{\hskip 2.5em} c @{\hskip 2.5em} c @{\hskip 2.5em} c}
\toprule
& \multicolumn{2}{c}{\textbf{Complexity per Step}} & \multicolumn{1}{c}{\textbf{Efficiency}} \\
\cmidrule(r{2.5em}){2-3} \cmidrule{4-4}
\textbf{Method} & \textbf{Time} & \textbf{VRAM} & \textbf{Forwards / Step} \\
\midrule
$\mu$Turn & $\mathcal{O}(N \cdot \mathbf{|Z|}^2)$ & $\mathcal{O}(\mathbf{|Z|}^2)$ & $N$ \\
MultiDiffusion &  $\mathcal{O}(\mathbf{|Z|}^2)\sim\mathcal{O}(N \cdot \mathbf{|Z|}^2)$ & $\mathcal{O}(N \cdot \mathbf{|Z|} + \mathbf{|Z|}^2)$ & $N$ \\
LoMOE & $\mathcal{O}(5N \cdot \mathbf{|Z|}^2)$ & $\mathcal{O}(N \cdot B \cdot \mathbf{|Z|}^2)$ & $\sim 5N$ \\
IDAttn & $\mathcal{O}((|\mathbf{Z}^\text{img}| + N \cdot |\mathbf{Z}^\text{prompt}_n|)^2)$ & $\mathcal{O}((|\mathbf{Z}^\text{img}| + N \cdot |\mathbf{Z}^\text{prompt}_n|)^2)$ & $1$ \\
{MICE} & $\mathcal{O}((|\mathbf{Z}^\text{img}| + N \cdot |\mathbf{Z}^\text{prompt}_n|)^2)$ & $\mathcal{O}((|\mathbf{Z}^\text{img}| + N \cdot |\mathbf{Z}^\text{prompt}_n|)^2)$ & $1$\\
\bottomrule
\end{tabular}%
}
\end{table*}

\section{Further Qualitative Comparison with the State-of-the-Art}
\begin{landscape}
    \begin{figure}
    \centering
    \includegraphics[width=\linewidth]{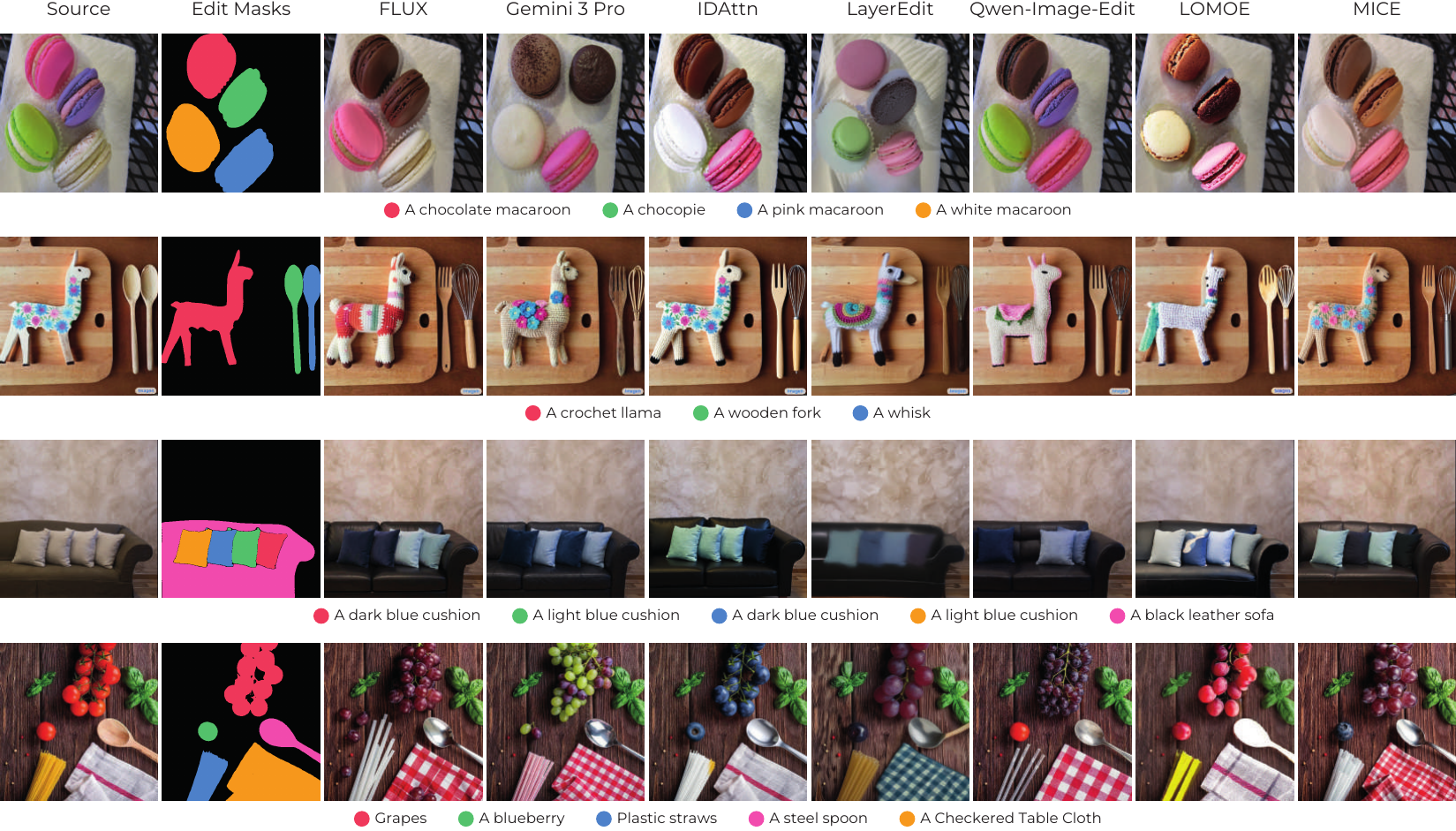}
    \caption{LoMOE-Bench leaderboard.}
    \label{fig:leaderboard_qualitatives_lomoe}
\end{figure}
\end{landscape}

\begin{landscape}
    \begin{figure}
    \centering
    \includegraphics[width=\linewidth]{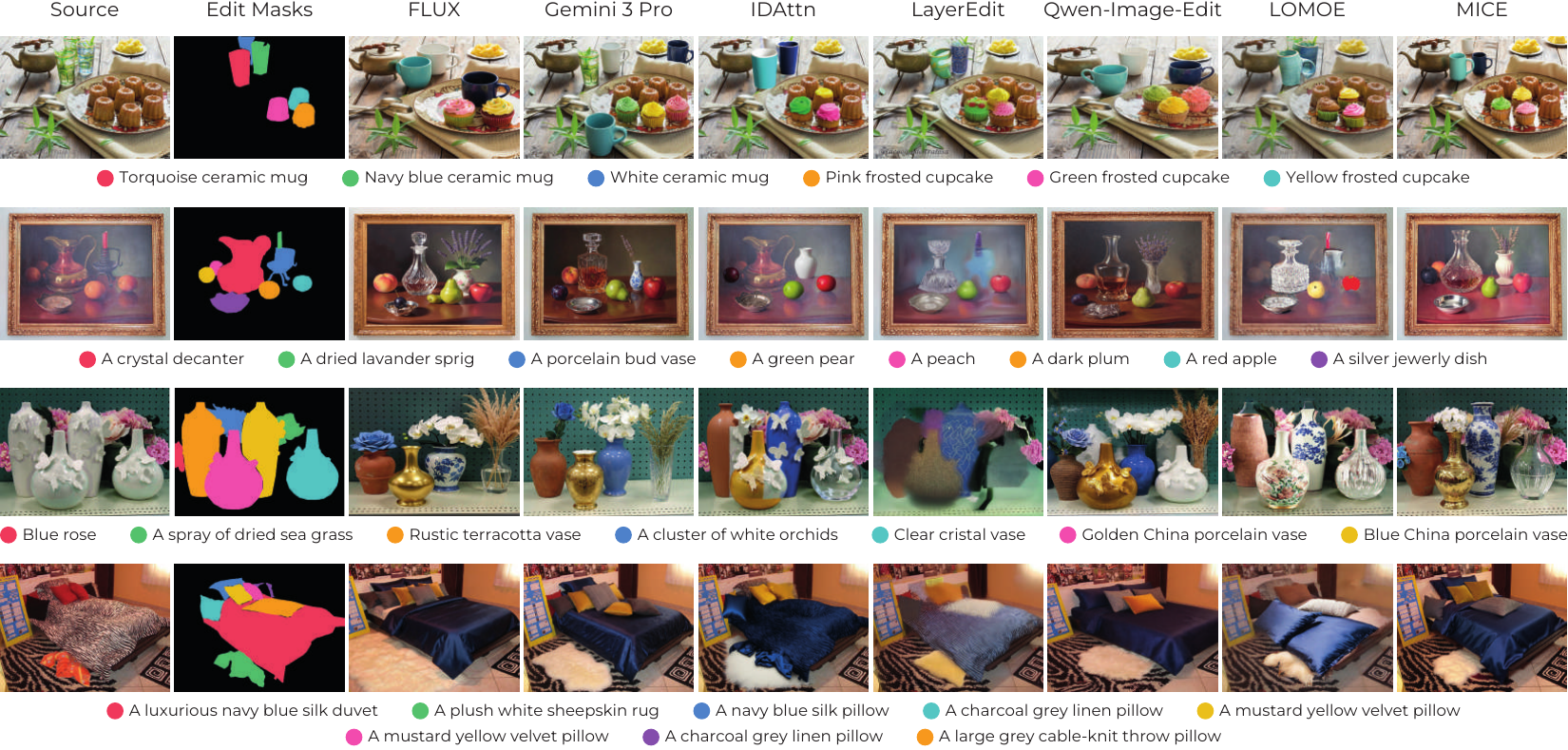}
    \caption{MICE-Bench leaderboard.}
    \label{fig:leaderboard_qualitatives_micebench}
\end{figure}
\end{landscape}

In this section, we show a broader qualitative comparison between our approach and the end-to-end baseline, the closed-source Gemini 3 Pro Image, IDAttn~\cite{zaccagnino2026shifting}, LayerEdit~\cite{fu2025layeredit}, LoMOE~\cite{goirik_lomoe}, and Qwen-Image-Edit~\cite{wu2025qwen_image}, on more samples from LoMOE-Bench (\Cref{fig:leaderboard_qualitatives_lomoe}) and MICE-Bench (\Cref{fig:leaderboard_qualitatives_micebench}).

\section{Exemplar Samples from MICE-Bench}
\begin{figure}
    \centering
    \includegraphics[width=\linewidth]{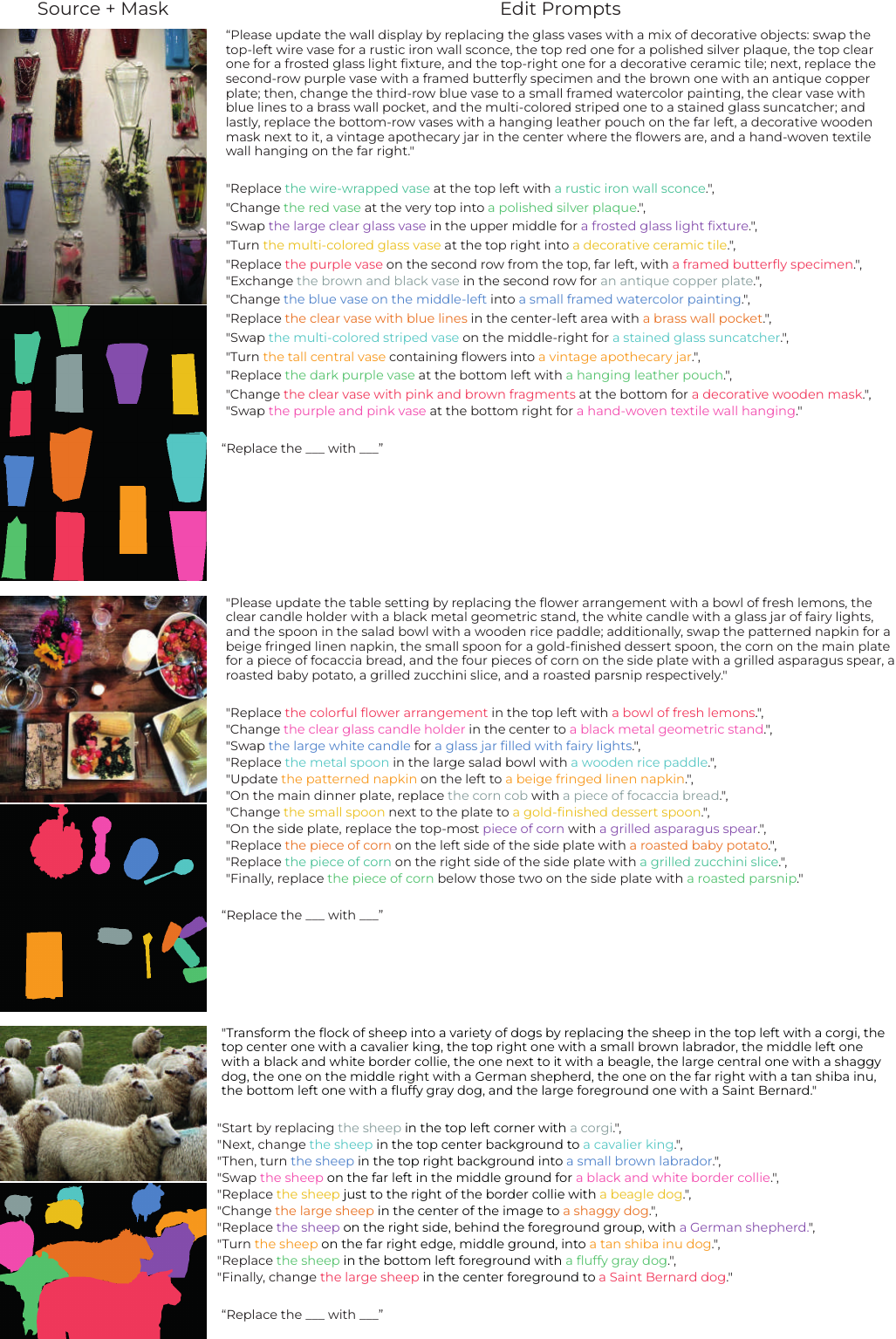}
    \caption{Some samples from the devised MICE-Bench dataset.}
    \label{fig:dataset_samples}
\end{figure}

We show samples for our proposed MICE-Bench in~\Cref{fig:dataset_samples}. For each source image, we provide the segmentation masks of the instances that should be edited. Moreover, we prepare three sets of textual prompts: 
\begin{itemize}
    \item A single, global prompt describing all the edits to perform and containing the localization information to identify each instance.
    \item A set of prompts, each with a self-contained editing instruction with localization information of the instance to edit.
    \item A set of programmatically-obtained prompts in the form of \texttt{Replace the \{SRC\} with \{TGT\}}, where \texttt{\{SRC\}} and \texttt{\{TGT\}} are the class of the instance to be edited and the description of the desired result, respectively.
\end{itemize}

\clearpage
\bibliographystyle{splncs04}
\bibliography{main}